\documentclass{article}

\PassOptionsToPackage{numbers, compress}{natbib}

\usepackage[final]{neurips_2022}

\usepackage[utf8]{inputenc} 
\usepackage[T1]{fontenc}    
\usepackage{hyperref}       
\usepackage{url}            
\usepackage{booktabs}       
\usepackage{amsfonts}       
\usepackage{nicefrac}       
\usepackage{microtype}      
\usepackage[dvipsnames]{xcolor}
\usepackage{microtype}
\usepackage{graphicx}
\usepackage{subfigure}
\usepackage{booktabs} 
\usepackage{algorithm}
\usepackage{algorithmic}
\usepackage{amsfonts}
\usepackage{amsmath}
\usepackage{amsthm}
\usepackage{bbm}
\usepackage{bbding}
\usepackage{wasysym}
\usepackage{pifont}
\newcommand{\cmark}{\text{\ding{51}}}

\usepackage{microtype}

\usepackage{graphicx}
\usepackage{subfigure}
\usepackage{booktabs} 
\usepackage{graphicx}
\usepackage{multirow}
\usepackage{amsfonts}
\usepackage{amsmath,amsfonts,amsthm,amssymb}
\usepackage{mathrsfs}
\usepackage{wrapfig}
\usepackage{microtype}
\usepackage{graphicx}
\usepackage{subfigure}
\usepackage{booktabs} 
\usepackage{algorithm}
\usepackage{algorithmic}
\usepackage{amsfonts}
\usepackage{amsmath}
\usepackage{amsthm}
\usepackage{bbm}
\usepackage{multirow}
\usepackage{bbding}
\usepackage{wasysym}
\usepackage{pifont}
\usepackage{tablefootnote}
\usepackage[font=small]{caption}
\usepackage{hyperref}
\usepackage{isomath}
\usepackage{mathtools}
\usepackage{setspace}
\usepackage{thm-restate}
\usepackage{url}
\usepackage{wrapfig}
\usepackage{enumitem}
\usepackage{hyperref}
\usepackage{amsmath}
\usepackage{amssymb}
\usepackage{mathtools}
\usepackage{amsthm}

\usepackage[capitalize,noabbrev]{cleveref}

\theoremstyle{plain}
\newtheorem{theorem}{Theorem}[section]
\newtheorem{proposition}[theorem]{Proposition}

\newtheorem{corollary}[theorem]{Corollary}
\theoremstyle{definition}

\declaretheorem[name=Remark]{reremark}
\usepackage[textsize=tiny]{todonotes}

\usepackage{tikz}

\title{Optimistic Curiosity Exploration and Conservative Exploitation with Linear Reward Shaping}

\author{%
  Hao Sun$^1$\thanks{hs789@cam.ac.uk. $^1$University of Cambridge; $^2$ Tencent RoboticsX; $^3$Hong Kong University of Science and Technology; $^4$ Tsinghua University; $^5$ IDEA; $^6$ University of California, Los Angeles} \\
   \And
   Lei Han$^2$ \\
    \And
   Rui Yang$^3$ \\
    \And
   Xiaoteng Ma$^4$ \\
    \And
   Jian Guo$^5$ \\
    \And
   Bolei Zhou$^6$ \\
}

\begin{document}

\maketitle

\begin{abstract}
In this work, we study the simple yet universally applicable case of reward shaping in value-based Deep Reinforcement Learning (DRL). We show that reward shifting in the form of a linear transformation is equivalent to changing the initialization of the $Q$-function in function approximation. Based on such an equivalence, we bring the key insight that a positive reward shifting leads to conservative exploitation, while a negative reward shifting leads to curiosity-driven exploration. Accordingly, conservative exploitation improves offline RL value estimation, and optimistic value estimation improves exploration for online RL. We validate our insight on a range of RL tasks and show its improvement over baselines: (1) In offline RL, the conservative exploitation leads to improved performance based on off-the-shelf algorithms; (2) In online continuous control, multiple value functions with different shifting constants can be used to tackle the exploration-exploitation dilemma for better sample efficiency; (3) In discrete control tasks, a negative reward shifting yields an improvement over the curiosity-based exploration method. 
\end{abstract}

\section{Introduction}
\begin{wrapfigure}{r}{6.5cm}
\vskip -0.17in
\centering
\begin{minipage}[htbp]{0.68\linewidth}
\centering
\includegraphics[width=1.0\textwidth]{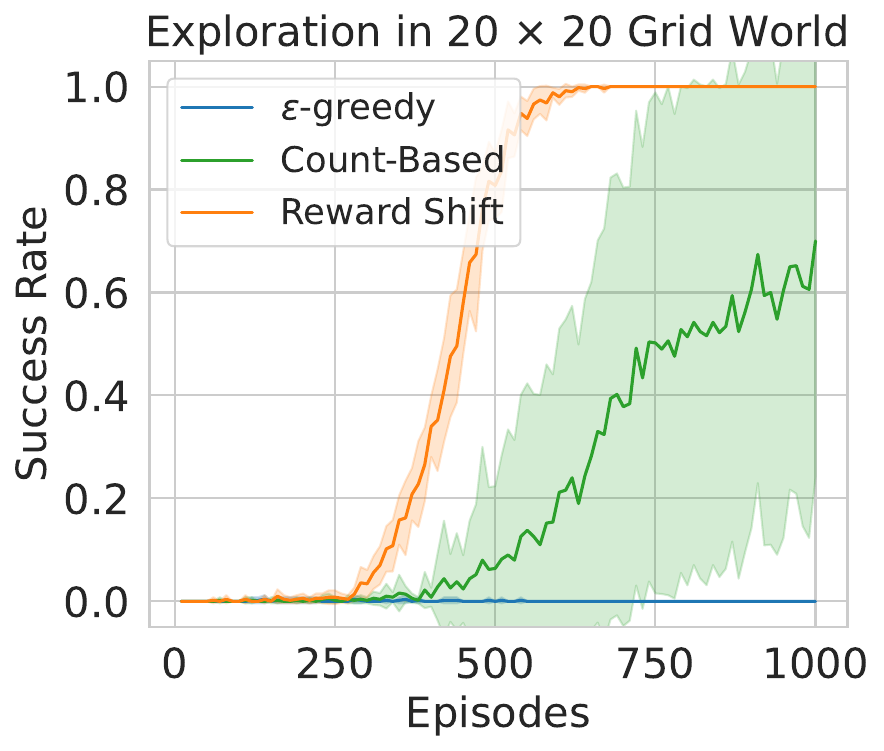}
\end{minipage}%
\begin{minipage}[htbp]{0.33\linewidth}
\includegraphics[width=0.95\textwidth]{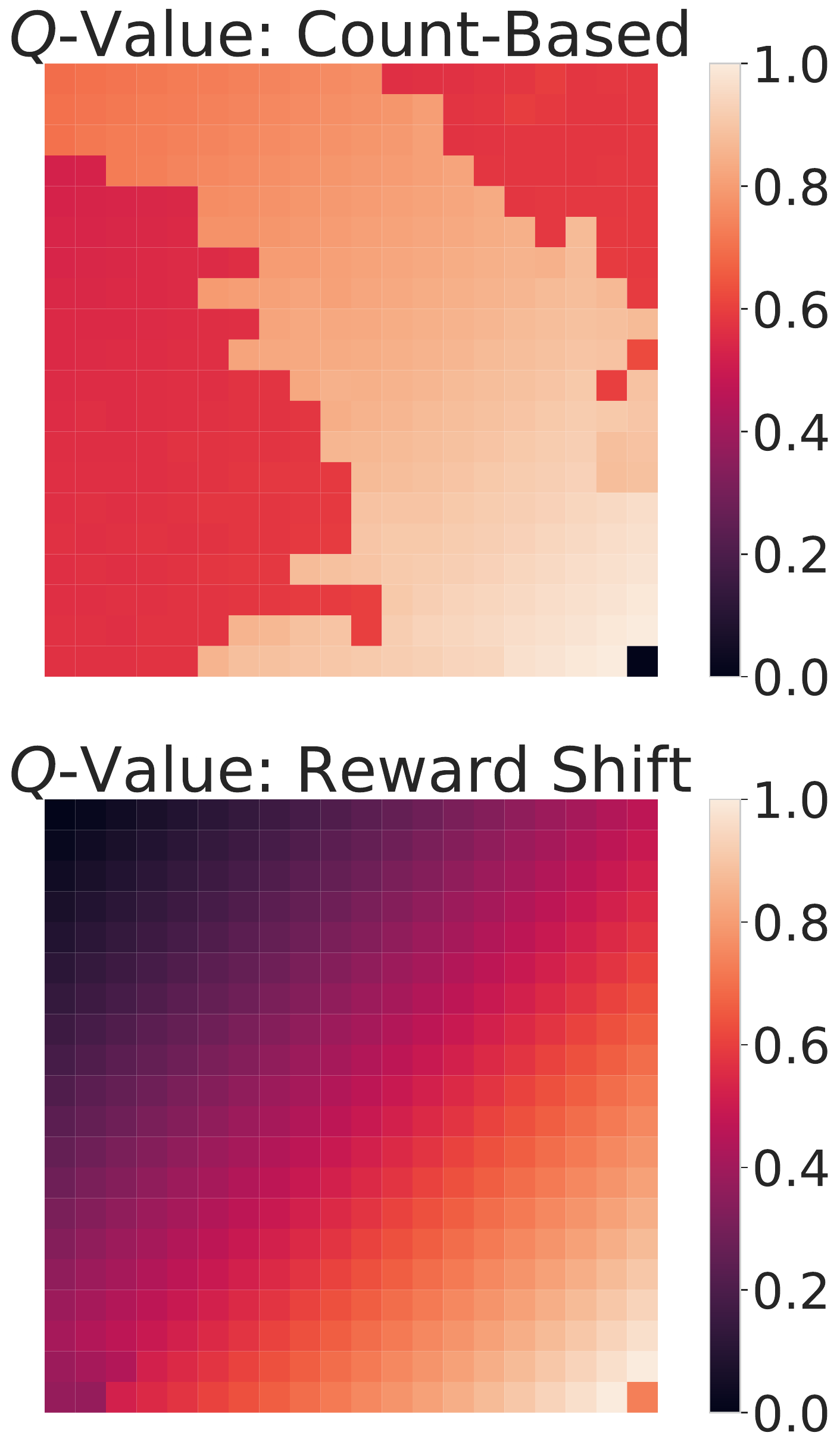}
\end{minipage}%
\vskip -0.1in
\caption{Our work is inspired by the observation that reward shifting remarkably helps exploration and outperforms count-based exploration in $Q$-learning (left). Reward shifting does not change the primal optimal $Q$-value landscape, and is able to learn a near-optimal $Q$-value (right). While count-based exploration suffers from the curse of dimensionality, reward shifting can be seamlessly applied to high-dim tasks including continuous control. 
}
\vskip -0.1in
\label{fig:motiv_demo}
\end{wrapfigure}


While reward shaping is a well-established practice in reinforcement learning and has a long-standing history~\citep{randlov1998learning,laud2004theory}, specifying a certain reward to incentivize the learning agent requires domain knowledge and a thorough understanding of the task~\citep{vinyals2019grandmaster,akkaya2019solving,berner2019dota,elbarbari2021ltlf}. Even with careful design and tuning, learning with a shaped reward that intends to accelerate learning may on the contrary hinder the learning performance by incurring the sub-optimal behaviors of the agent~\citep{florensa2017reverse,plappert2018multi}. Although \citet{ng1999policy} theoretically points out that optimal policy remains unchanged under certain form of reward transformation, and in the later work of \citet{wiewiora2003principled} a framework is proposed to guide policies with prior knowledge under tabular setting, how reward shifting accommodates recent Deep Reinforcement Learning (DRL) algorithms remains much less explored.

In this work, we study a special linear transformation, which is the simplest form of reward shaping, in value-based DRL~\citep{sutton1998reinforcement,lillicrap2015continuous,mnih2015human,fujimoto2018addressing}. We start with understanding how such a specific kind of reward shaping works in value-based DRL function approximations and show that reward shifting is equivalent to different $Q$-value initialization, extending previous discovery of ~\citep{wiewiora2003principled} to the function approximation setting. Figure~\ref{fig:motiv_demo} showcases reward shifting benefits exploration in a maze game and outperforms count-based and $\epsilon$-greedy exploration and learns near optimal value function.

Based on such an equivalence, we introduce the key insight of this work:

\textbf{\emph{~~~~A positive reward shifting leads to conservative exploitation, while a negative reward shifting leads to curiosity-driven exploration. }}

We demonstrate the application of such an insight in \textcolor{purple}{\textbf{three Deep-RL scenarios:}} \textcolor{purple}{\textbf{(S1) for offline RL}}, we show that conservative exploitation induced by reward shifting improves learning performance of off-the-shelf algorithms; \textcolor{purple}{\textbf{(S2) for online RL setting}}, we show multiple value functions with different reward shifting constants can be used to trade-off between exploration and exploitation, thus improve learning efficiency; \textcolor{purple}{\textbf{(S3) for curiosity-driven exploration,}} we introduce a simple yet crucial adaptation on a prevailing curiosity-based exploration algorithm, the Random Network Distillation~\citep{burda2018exploration}, making it compatible with value-based DRL algorithms.
Experiments on a diverse set of tasks including continuous and discrete action space control show our method brings substantial improvement over baselines.




\textbf{Our contributions} can be summarized as follows:
\begin{enumerate}[nolistsep]
    \item Analytically, we introduce the key insight that reward shifting is equivalent to diversified $Q$-value network initialization in value-based DRL, which can be applied to curiosity-driven exploration and conservative exploitation;
    \item Practically, we instantiate the key insight to three different scenarios, namely the offline conservative exploitation, sample-efficient continuous control, and curiosity-driven exploration, to contrast the generality of reward shifting;
    \item Empirically, we demonstrate the effectiveness of the proposed method integrated with multiple off-the-shelf baselines on both continuous and discrete control tasks. 
\end{enumerate}

\section{Preliminaries and Related Work}
\subsection{Online RL}
We follow a standard MDP formulation in the online RL setting, i.e., $\mathcal{M} = \{\mathcal{S},\mathcal{A},\mathcal{T},\mathcal{R},\rho_0,\gamma,T\}$, where $\mathcal{S}\subset \mathbb{R}^{d}$ denotes the $d$-dim state space, $\mathcal{A}$ is the action space (note for discrete action space $|\mathcal{A}|<\infty$ and for continuous control $|\mathcal{A}|=\infty$), we consider a deterministic transition dynamics $\mathcal{T}: \mathcal{S}\times \mathcal{A} \mapsto \mathcal{S}$ and deterministic reward function $\mathcal{R}:\mathcal{S}\times\mathcal{A}\mapsto \mathbb{R}$. $\rho_0 = p(s_0)\in\Delta(\mathcal{S})$ denotes the initial state distribution. $\gamma$ is the discount factor and $T$ is the episodic decision horizon.
Online RL considers the problem of learning a policy $\pi\in\Pi:\mathcal{S}\mapsto \Delta (\mathcal{A})$ (or $\pi\in\Pi:\mathcal{S}\mapsto\mathcal{A}$ with a deterministic policy class), such that the expected cumulative reward $\mathbb{E}_{a_t\sim\pi, s_{t+1}\sim \mathcal{T},s_0\sim \rho_0}\sum_{t=0}^T \gamma^t r_t(s_t, a_t)$ in the Markovian decision process is maximized.
In online RL setting, an agent learns through trials and errors~\citep{sutton1998reinforcement}, through either on-policy~\citep{schulman2015trust,schulman2017proximal,cobbe2021phasic} or off-policy paradigm~\citep{lillicrap2015continuous,mnih2015human,fujimoto2018addressing,wang2016sample,haarnoja2018soft,sun2020zeroth}. 
In this work, we focus on the off-policy value-based methods which are in general more sample efficient. Specifically, our discussions assume the policy learning is based on a learned $Q$-value function that approximates the cumulative reward an agent can gain in the following part of an episode. Formally, $Q(s_t,a_t) = \mathbb{E}_{\pi,\mathcal{T}}\sum_{\tau=t}^T\gamma^t r(s_\tau, a_\tau)$, and can be estimated by propagating the Bellman operator $\mathbb{B}Q(s,a) = r(s,a) + \gamma \mathbb{E}Q(s',a')$. For value-based methods, the (soft-)optimal policy is then produced by 
\begin{equation}
    \pi^*_\alpha(a|s) = \frac{\exp \frac{1}{\alpha}Q^*(s,a)}{\sum_{a'} \exp \frac{1}{\alpha}Q^*(s,a')},
\end{equation} 
where $Q^*$ is the optimal $Q$-value function. We can also set the temperature parameter close to 0 to have the deterministic policy class. Simplifying the notion we have $\pi^*(s) = \arg\max_a Q^*(s,a)$. Algorithms like DPG~\citep{silver2014deterministic} can be used to address the intractable analytical argmax issue that arises in continuous action space. We choose to develop our work on top of prevailing baseline algorithms of DQN~\citep{mnih2015human}, BCQ~\citep{fujimoto2018off}, CQL~\citep{bharadhwaj2020conservative} and TD3~\citep{fujimoto2018addressing} as a minimal example to isolate the source of gains. It should be straightforward to extend the method on top of other learning algorithms. 

\subsection{Exploration and the Curiosity-Driven Methods}
\begin{table*}[t]
	\centering
	\fontsize{6.5}{6.5}\selectfont 
	\caption{Reward shifting is flexible to be plugged into both online and offline RL algorithms to guarantee conservative exploitation or pursue optimistic exploration. It covers both discrete and continuous control tasks, with only a little additional computational expense. Moreover, the optimal policy learned with shifted reward is not biased.}
	\begin{tabular}{ccccccccc}
		\toprule
		Covered Topics &\multicolumn{1}{c}{\multirow{1}{*}{Related Work}} & Plug-in & Online & Offline & Discrete & Continuous & Unbiased & Examples \cr
		\cmidrule(lr){1-2}\cmidrule(lr){3-9}
		\multicolumn{1}{c}{\multirow{4}{*}{Exploration}}
    	 & Curiosity &$\cmark$ & $\cmark$ & $\cdot$ & $\cmark$ & $\cdot$ & $\cdot$ &  \citet{burda2018exploration} \cr
    	 & Ensemble &$\cdot$ & $\cmark$ & $\cdot$ & $\cmark$ & $\cdot$ & $\cmark$ & \citet{osband2018randomized} \cr
    	 & Initialization &$\cdot$ &  $\cmark$ & $\cdot$ & $\cmark$ & $\cdot$& $\cmark$  & \citet{rashid2020optimistic} \cr
    	 & Optimism &$\cdot$ &$\cmark$ &  $\cdot$ & $\cdot$ & $\cmark$& $\cdot$ & \citet{ciosek2019better} \cr
		\cmidrule(lr){1-2}\cmidrule(lr){3-9}
		\multicolumn{1}{c}{\multirow{2}{*}{Exploitation}}
    	 & Conservatism &$\cdot$ & $\cdot$ & $\cmark$ & $\cmark$ & $\cmark$ & $\cdot$ & \citet{bharadhwaj2020conservative} \cr
    	 & Policy Constraints &$\cmark$ & $\cdot$ & $\cmark$ & $\cmark$ & $\cmark$ & $\cmark$ & \citet{fujimoto2018off} \cr
		\cmidrule(lr){1-9}
    	 & Reward Shifting & $\cmark$ & $\cmark$ & $\cmark$ & $\cmark$ & $\cmark$ & $\cmark$& Ours \cr
		\bottomrule
	\end{tabular}\vspace{-0.4cm}
	\label{tab:relatedwork}
\end{table*}

One of the most important issues in online RL is the exploration-exploitation dilemma~\citep{sutton1998reinforcement} that the agent must simultaneously learn to exploit its accumulated knowledge on the task and explore new states and actions. Previous works address the exploration problem from various perspectives: 
for discrete action space tasks, count-based methods like \citep{bellemare2016unifying,ostrovski2017count,tang2017exploration} are proposed to motivate the revisiting of under-explored state-action pairs. Specifically, \citet{choshen2018dora} extended the idea into general settings by constructing an additional MDP for Exploration-value estimation, as a generalized counter for count-based exploration. 
To boost exploration in continuous state tasks, curiosity-driven methods are investigated by
\citet{houthooft2016variational,pathak2017curiosity,burda2018large,burda2018exploration}, where variety of intrinsic rewards are designed as supplementary to the primal task reward for better exploration. Self-imitate approaches like ~\citet{oh2018self,ecoffet2019go,sun2019policy} repeat success trajectories but require extra assumptions on the environment.

DORA~\citep{choshen2018dora} constructed an additional MDP to estimate the Exploration-value as a generalised counter for count-based exploration, yet those count-based methods are orthogonal to reward shifting: in intrinsic reward methods, an agent must \textbf{first experience} a new $(s,a)$ pair before receiving a high intrinsic reward --- this is extremely hard with an arg-max style policy. On the other hand, with optimistic initialization, the rarely-visited $(s,a)$ pairs will naturally have higher $Q$-values \textbf{before experiencing} it --- as the frequently-visited pairs have updated their values with a negatively shifted reward. From such a perspective, reward shifting not only works by itself motivates exploratory behaviors, but can also be seamlessly plugged into intrinsic reward methods to \textbf{encourage the first visitation} of new states.

The works of DIAYN and DADS~\citep{eysenbach2018diversity,sharma2019dynamics} show that various skills can be developed even without the primal extrinsic reward. For continuous control tasks, OAC~\citep{ciosek2019better} improves the SAC~\citep{haarnoja2018soft} with informative action space noise based on the optimism in face of uncertainty (OFU)~\citep{brafman2002r,jaksch2010near,azar2017minimax,jin2018q}. GAC~\citep{tessler2019distributional} addresses the exploration issue with a richer functional class for the policy.

In the work of \citet{rashid2020optimistic}, the problematic pessimistic initialization is addressed for better exploration, yet the work focuses on specific settings of tabular and discrete control exploration. In the work of \citet{osband2016deep,osband2018randomized}, ensemble models with diverse initialization and randomized priors are used to resemble the insight of bootstrap sampling and facilitate better value estimation, yet those methods are only applicable to discrete control tasks. Noted that although the reward shifting can be regarded as a special case of these random priors, it can be distinguished by not changing the optimal $Q$-value, and its flexibility to be plugged into both continuous and discrete control algorithms.

Random Network Distillation (RND)~\citep{burda2018exploration} defines the  intrinsic reward as the output difference between a fixed neural network $\phi_1$ and a trainable network $\phi_2$ given state-actions as the inputs. e.g., 
\begin{equation}
    r_{\mathrm{int}}(s,a) = |\phi_2(s,a) - \phi_1(s,a)|,
\end{equation}
where both networks are activated by a sigmoid function. After optimizing the learnable $\phi_2$ to approximate $\phi_1$ with seen $(s,a)$ pairs~\footnote{Or computed with only states, i.e., $r_{\mathrm{int}}(s) = |\phi_2(s) - \phi_1(s)|$.}, the value of $r_{\mathrm{int}}(s,a)$ will decay to $0$ for such state-action pairs that are frequently visited but remain high for thoase are seldom visited.

In this work, we show that exploratory behavior can be achieved simply by shifting the reward function with a constant. Thus our method is orthogonal to those previous approaches in the sense that our intrinsic exploration behavior is motivated by high function approximation error in the under-explored state-action pairs. We demonstrate such insight by showing that RND in its original design is not suitable for value-based methods in developing exploratory behaviors, but integrating RND with a shifted reward function remarkably improves the learning performance.

\subsection{Offline RL}
The offline RL, also known as batch-RL, focuses on the problems where interaction with the environment is impossible, and the policy can only be optimized based on the logged dataset. In those tasks, a fixed buffer $\mathcal{B} = \{s_i,a_i,r_i,s_i'\}_{i=[N]}$, collected from some unknown behavior policy $\pi_\beta$, is provided. In general, such a dataset can either be generated by rolling out an expert that generates high-quality solutions to the task~\citep{fujimoto2018off,zhang2020brac+,fu2020d4rl} or a non-expert that executes sub-optimal behaviors~\citep{fu2020d4rl,wu2019behavior, kumar2019stabilizing, agarwal2020optimistic, jarrett2021inverse} or be a mixture of both~\citep{bharadhwaj2020conservative}. As the agent in the offline RL setting can not correct its potentially biased knowledge through interactions, the most important issue is to address the extrapolation error~\citep{fujimoto2018off} induced by distributional mismatch~\citep{kumar2019stabilizing}. To address such an issue, a series of algorithms optimize the policy learning under the constraint of distributional similarity~\citep{wu2019behavior,kumar2019stabilizing,siegel2020keep,yang2022rorl}. 

\citet{bharadhwaj2020conservative} propose CQL to solve the offline RL tasks with a conservative value estimation. Specifically, CQL learns the $Q$-value estimation by jointly maximizing the $Q$-values of actions sampled from the behavior offline dataset and minimizing the $Q$-values of actions sampled with pre-defined prior distributions (e.g., uniform distribution over the action space). As we will show in this work, an alternative approach to have a lower bound for the optimal $Q$-value function is to use an appropriately shifted reward function. This idea leads to the direct application of our proposed framework in the offline setting. In general, reward shift can be plugged in many distribution-matching offline-RL algorithms~\citep{fujimoto2018off,wu2019behavior,kumar2019stabilizing,siegel2020keep} to further improve the performance with conservative $Q$-value estimation.

Table~\ref{tab:relatedwork} contextualizes reward shifting with respect to related works we discussed above.

\section{A Motivating Example}
We start with two intuitive remarks and a ``counter-intuitive'' motivating example in this section before introducing our method. 

\begin{reremark}
\label{remark_1}
Given an MDP $\mathcal{M} = \{\mathcal{S},\mathcal{A},\mathcal{T},\mathcal{R},\rho_0,\gamma\}$, where $|\mathcal{A}| < \infty$, scaling the reward function with linear transformation, i.e., $\mathcal{R}_{k,b} = k\cdot\mathcal{R} + b$, $\forall k>0, b\in \mathbb{R}$, such that $r'_t = kr_t+b \in \mathcal{R}_{k,b}$, does not change the optimal policy induced by the corresponding value function $Q^*_{k,b}(s,a):=\sum_t
\gamma^t r'_{t}$:
\begin{small}
\begin{equation}
\begin{split}
    \pi^*(s) = \arg \max_{a\in \mathcal{A}} Q^*_{k,b}(s,a) 
    = \arg \max_{a\in \mathcal{A}} kQ^*(s,a) + \frac{b}{1-\gamma} 
    = \arg \max_{a\in \mathcal{A}} Q^*(s,a),
\end{split}
\end{equation}
\end{small}
\end{reremark}
\begin{reremark}
\label{remark_2}
When $|\mathcal{A}|=\infty$, scaling the reward function with linear transformation does not change the optimal policy induced by deterministic policy gradient~\citep{silver2014deterministic}, given proper learning rate $\eta' = \eta/k $:
\begin{small}
\begin{equation}
\begin{split}
    \nabla_\theta J(\mu_\theta) &= \mathbb{E}_{s_t}[\nabla_{a} Q^*(s_t, a_t)|_{a_t=\mu_{\theta}(s_t)} \nabla_\theta\mu_{\theta}(s_t)] =  \mathbb{E}_{s_t}[\nabla_{a} Q_{k,b}^*(s_t, a_t)|_{a_t=\mu_{\theta}(s_t)} \nabla_\theta\mu_{\theta}(s_t)]/k,
\end{split}
\end{equation}
\end{small}
\end{reremark}
Remark~\ref{remark_1} and Remark~\ref{remark_2} declare the fact that constant reward shifting does not affect the optimal policy induced by the optimal $Q$-value function calculated by the shifted reward. However, Figure~\ref{fig:motiv_demo} presents ``counter-intuitive'' results in a demonstrative exploration task of Grid World. In this task, an agent located at the upper left corner of a map needs to explore without reward and reach the goal point located at the lower right corner. A non-trivial reward of $+1$ will be assigned only when the goal is reached. We report learning curves with regard to the episodic success rate in reaching the goal point and the learned $Q$-values. In this toy example, we find a negative reward shifting remarkably boosts the learning efficiency of $Q$-learning, and surpasses the conventional count-based method for exploration. 
Moreover, Remark~\ref{remark_1} is empirically verified with such a toy example: reward shifting does not change the optimal $Q$-value as well as its induced policy, but on the contrary accelerates the discovery of (near-)optimal policy.
\begin{figure*}[t!]
    \centering
    \vspace{0.5cm}
    \includegraphics[width=1.02\textwidth]{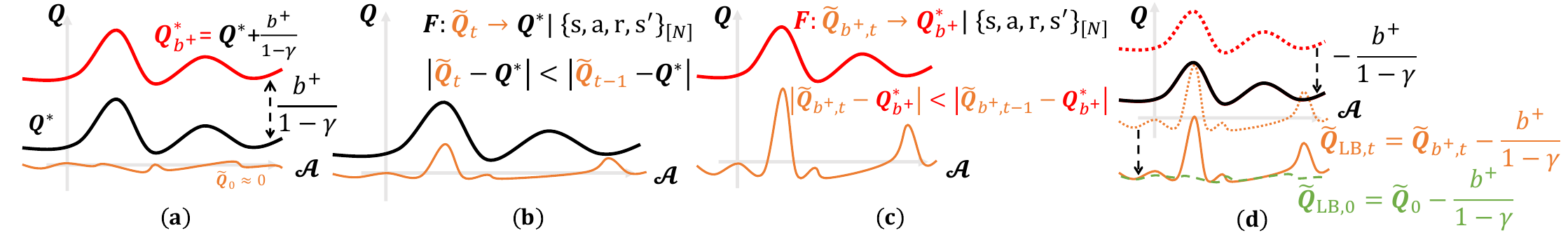}
    \vspace{-0.5cm}
    \caption{Illustrative figure for conservative exploitation, with a positive constant bias added to the reward function. 
    We use \textcolor{black}{\textbf{Black}} lines to denote the original value function, and use  \textcolor{red}{\textbf{Red}}  lines to denote the shifted value function with the constant reward shift. \textcolor{orange}{\textbf{Orange}} lines are used to denote the function approximation during different learning stages and \textcolor{Green}{\textbf{Green}} line shows the equivalent \textbf{pessimistic initialization}.
    \textbf{(a)} shifting the reward function with a positive bias term $b^+$ leads to an uniformly increased $Q$-value function, namely $\textcolor{red}{Q_{b^+}^*} = Q^* + \frac{b^+}{1-\gamma}$, during learning, a neural network estimator \textcolor{orange}{$\tilde{Q}$} initialized with $\textcolor{orange}{\tilde{Q}_0}\approx0$ is optimized to approximate the $Q$-value functions (e.g., through TD or MC estimation).
\textbf{(b)} for any value-based RL algorithm, the value optimization step can be regarded as a function $F$ that minimizes the difference between the estimated $Q$-value function \textcolor{orange}{$\tilde{Q}_t$} and the optimal one $Q^*$, given the interaction experience with the environment (e.g., a replay buffer $\mathcal{B}$ for off-policy methods). Note \textcolor{orange}{$\tilde{Q}_t$} approximating $Q^*$ better than \textcolor{orange}{$\tilde{Q}_{t - 1}$} holds in expectation as long as the RL algorithm is designed to approximate the optimal value function.
\textbf{(c)} similarly, the optimizer given the same interactive experience (e.g., replay buffer $\mathcal{B}$) will learn to minimize the difference between $Q$-value function \textcolor{orange}{$\tilde{Q}_{b^+,t}$} and the optimal one \textcolor{red}{$Q_{b^+}^*$}, after re-labeling the rewards in the buffer by $r' = r + b^+$.
\textbf{(d)} according to Remark~\ref{remark_2}, the optimization conducted in (c) is equivalent to (b) with the neural network $Q$-value estimator initialized as $\textcolor{Green}{\tilde{Q}_{\mathrm{LB},0}}\approx \textcolor{orange}{\tilde{Q}_0} - \frac{b^+}{1-\gamma}$, rather than $\textcolor{orange}{\tilde{Q}_0}\approx0$. i.e., by shifting the reward with proper positive value $b^+$, we are able to initialize the $Q$-value network that lower-bounds the optimal $Q$-value.}
    \label{fig:OCEAN}
\end{figure*}

In the following of this work, we study the effect of varying constant bias $b$ and fix the scaling factor $k=1$ to avoid trivial discussions on the choices of learning rate --- there should be no surprise that choosing an appropriate learning rate is empirically important. 
We focus on revealing the importance of selecting the universal bias term $b$ in the reward function through the lens of initialization priors in function approximation and show such a bias is generally helpful for both online and offline settings. In the online settings, it improves learning efficiency, and in the offline settings, it promotes conservative exploitation.


\section{Shifted Priors for $Q$-value Estimation}
\subsection{Reward Shift Equals to Different Initialization}
\label{sec:method_1}

We start by formally introducing notions and the key idea of this work: reward shifting equals different initialization. 
We use Figure~\ref{fig:OCEAN} to illustrate how reward shifting affects function approximation, hence changing the learning dynamics for value-based algorithms.
The original optimal $Q$-value function is denoted as $Q^*$, and plotted in the figures as \textcolor{black}{\textbf{Black}} 
curves. We then denote the shifted optimal $Q$-value function as $Q^*_{b^+}$, which is the $Q$-value function with the shifted reward $r'=r+b^+$. We use \textcolor{red}{\textbf{Red}} curves in figures to denote those shifted $Q$-value functions. In this section we provide analysis with a positive bias $b^+>0$ to illustrate how positively shifted value function affects function approximation, and coordinates the normally applied near-zero initialized function approximators (e.g., neural networks. \textcolor{orange}{\textbf{Orange }} curves in the figure) to inspire conservative behaviors. The discussion of negative biases is elaborated in the next section.

To summarize, shifting the reward function with a positive constant is equivalent to initializing the value network with a smaller value --- as the $Q$-value of unseen state-action pairs during training are much lower than their shifted optimal values, those actions will not be selected in argmax-style policy updates --- leading to conservative learning behaviors that benefit policy learning in offline settings.

\subsection{\textcolor{purple}{\textbf{(S1) Offline RL}}: Conservative Exploitation}
\label{sec:offline}
According to the key insight we presented in Section~\ref{sec:method_1}, conservative $Q$-value estimation can emerge with a positively shifted reward. And such a value estimation empirically lower-bounds the optimal $Q$-value function.
\begin{figure*}[t]
    \centering
    \vspace{-0.1cm}
    \includegraphics[width=0.9\textwidth]{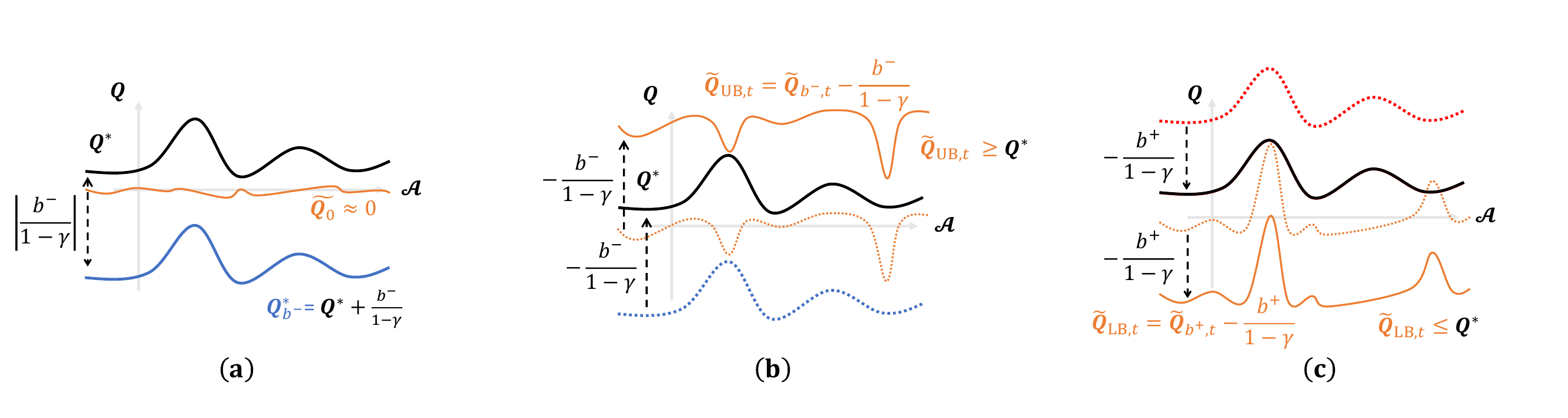}
    \vspace{-0.1cm}
    \caption{Illustrative figure for curiosity-driven exploration with a negative shifted reward. \textbf{(a)} While adding a negative constant value $b^{-}$ on the reward function leads to negatively shifted optimal $Q$-value function \textcolor{NavyBlue}{$Q_{b^-}^*$}. \textbf{(b)} Minimizing the difference between a $Q$-value approximator and the optimal $Q$-value will enable calculating an upper-bound estimation for $Q^*$ which can be used for \textbf{optimistic exploration}. \textbf{(c)} A positive constant shift added to the reward function can be used for \textbf{conservative policy update}.}
    \label{fig:OCEAN_2}
    \vspace{-0.3cm}
\end{figure*}
As shown in Figure~\ref{fig:OCEAN}(a), a positive constant $b^+$ added to the reward function will lead to a universal positively shifted optimal $Q$-value function, and the gap between the primal $Q$-value function and the new one is $\frac{b^+}{1-\gamma}$. Optimizing the  $Q$-value approximator with logged data will minimize the difference between the predicted value and the optimal value with observed data. For the unobserved data, the pessimistically initialized $Q$-value approximator guarantees the prediction is lower than the positively shifted optimal ones, thus conservative exploitation can be achieved with argmax-style value propagation and action execution on such a value function.

\subsection{(Curiosity-Driven) Optimistic Exploration}

On the other hand, if we shift the reward function to the negative side, it is equivalent to an optimistic initialization. Figure~\ref{fig:OCEAN_2} (a-b) illustrate how adding a negative bias leads to curiosity-driven exploration. With sufficiently small $b^-$ (so that $|b^-|$ larger than the maximal value of any $(s,a)$-pair), such an upper bound of $Q^*$ can be used to conduct curiosity-driven exploration. Intuitively, initializing a value network that always predicts a value larger than the true optimal value will lead to curiosity-driven exploration behavior, as any visited state will be assigned a relatively smaller value and the argmax-style policy will tend to choose under-explored actions.

Based on the analysis above that (1) a positive constant shift added to the reward function can be used for conservative policy updates and (2) a negative constant shift added to the reward function can be used for curiosity-driven exploration, we are ready to access both the upper bound, i.e., the optimistic estimation with $b^-$, and the lower bound, i.e., the conservative estimation with $b^+$ of the optimal value function. Formally, we use 
\begin{small}
\begin{equation}
    \tilde{Q}_{\mathrm{LB},t}(s,a) = \tilde{Q}_{b^+,t}(s,a) - \frac{b^+}{1-\gamma}
\end{equation}
\end{small}
to denote the empirical lower bound of value estimation (cf. Figure~\ref{fig:OCEAN}(d), \textcolor{orange}{\textbf{Orange}} line), and 
\begin{small}
\begin{equation}
\tilde{Q}_{\mathrm{UB},t}(s,a) = \tilde{Q}_{b^-,t}(s,a) - \frac{b^-}{1-\gamma}
\end{equation}
\end{small}
to denote the empirical upper bound of value estimation (cf. Figure~\ref{fig:OCEAN_2}(b), \textcolor{orange}{\textbf{Orange}} line). In both notions, $t$ denotes the optimization step. When $t=0$, with a near-zero initialization $\tilde{Q}_{0}\approx 0$, $\tilde{Q}_{\mathrm{LB},0}(s,a) = - \frac{b^+}{1-\gamma}$ is able to lower bound the unknown optimal $Q$-value given sufficiently large $b^+$. (cf. Figure~\ref{fig:OCEAN}(d), \textcolor{Green}{\textbf{Green}} line). Similarly, $Q$-value is upper-bounded by $\tilde{Q}_{\mathrm{UB},0}(s,a)$ with a sufficiently small $b^-$. 

Following those notions, we introduce our sample-efficient algorithms for both continuous control and discrete action space respectively. We propose a practical algorithm for sample-efficient continuous control in Section~\ref{sec:continuous_control}, and focus on a special class of curiosity-driven exploration method, the RND~\citep{burda2018exploration}, in Section~\ref{sec:curiosity_driven}.

\subsubsection{\textcolor{purple}{\textbf{(S2) Online RL}}: Trading-off Exploration and Exploitation with Reward Shift}
\label{sec:continuous_control}
According to the principle of optimism in the face of uncertainty (OFU), an exploration bonus that manifests the uncertainty of the $Q$-value function can be introduced into the value estimation: integrating optimistic exploration with conservative exploitation,
\begin{small}
\begin{equation}
\label{eqn:ofu4}
\begin{split}
    \hat{Q}(s,a) & = \tilde{Q}_{\mathrm{LB},t}(s,a) + \beta [\tilde{Q}_{\mathrm{UB},t}(s,a) - \tilde{Q}_{\mathrm{LB},t}(s,a) ] \\
    & = (1-\beta)(\tilde{Q}_{b^+,t}(s,a) - \frac{b^+}{1-\gamma} )
    + \beta (\tilde{Q}_{b^-,t}(s,a) - \frac{b^-}{1-\gamma}) \\
    & = (1-\beta)\tilde{Q}_{b^+,t}(s,a) + \beta \tilde{Q}_{b^-,t}(s,a)
    - \frac{(1-\beta) b^+ + \beta b^-}{1-\gamma},
\end{split}
\end{equation}
\end{small}
where the second term with coefficient $\beta$ denotes exploration bonus that is composed of uncertainty.

For those under-explored state-action pairs, i.e., extremely out-of-distribution samples for our neural network, both $\tilde{Q}_{b^+,t}(s,a)$ and $\tilde{Q}_{b^-,t}(s,a)$ will give near-zero predictions as a consequence of initialization (detailed implications are provided in Appendix~\ref{appdx:implication}).
Hence, the explorative bonus becomes $-\frac{(1-\beta)b^+ + \beta b^-}{1-\gamma}$, which is equivalent to applying another constant reward shift with value of $c_r = (1-\beta)b^+ + \beta b^-$, formally, we have
\begin{proposition}
\label{prop1}
Assuming we have access to an unbiased estimator for the optimal value function $Q^*$, e.g., with Monte-Carlo estimation $\hat{Q}^* = \mathbb{E}\sum_{t}\gamma^t r$, and the optimization is based on minimizing the MSE between the unbiased estimator and the function approximator, i.e., $\epsilon_t^2 = (\tilde{Q}_{t} - \hat{Q}^*)^2$, $\tilde{Q}_{t} = \tilde{Q}_{t-1} - 2\eta(\tilde{Q}_{t-1} - \hat{Q}^*)$, then the linear combination in Equation (\ref{eqn:ofu4}) is equivalent to a linear combination of the constants with value of $c_r = (1-\beta)b^+ + \beta b^-$.
\end{proposition}
The proof can be found in the Appendix \ref{appdx:proof}. According to Proposition~\ref{prop1}, a grid search for trading-off between the three hyper-parameters, i.e., the exploration bias $b^-$, the exploitation bias $b^+$ and the coefficient $\beta$ in Equation (\ref{eqn:ofu4}) is trivial as they only lead to a linear combination as $c_r = (1-\beta)b^+ + \beta b^-$, indicating that
\begin{corollary}
Changing the reward shifting constant $b$ is sufficient to trade off between exploration and exploitation.
\end{corollary}
The corollary says, in principle, a meta-learner can be trained to monitor the learning process and select a proper reward shifting constant automatically~\citep{duan2016rl,portelas2020teacher} to balance exploration and exploitation. For the ease of exposition, in this work we choose to focus on the simplest yet effective uniform sampling strategy from multiple shift constants, which has been shown as a strong baseline of those meta-learner approaches~\citep{graves2017automated,matiisen2019teacher}, and leave more complicated meta-learner-based shifting constant adjustment for future investigation.

Specifically, we use multiple Q-networks to learn with transition tuples $(s, a, r, s')$ sampled from the identical buffer that collects the policy's historcal interactions with the environment. In propagating Q-values through the temporal difference loss, the primal recorded reward value $r$ is replaced with shifted rewards with \textit{different} constant biases to update their individual Q-networks. We then uniformly sample one of those learned Q-networks for the optimization of policy networks (e.g., with  DPG~\citep{silver2014deterministic}). It is worth noting that our approach only requires post-hoc revision of the primal reward function, rather than interacting with the environment multiple times to collect samples for each value network. We dub the proposed method Random Reward Shift (RRS), and provide the pseudo-code in Algorithm~\ref{algo_RRS} of Appendix~\ref{appdx:cc_detail}.

\subsubsection{\textcolor{purple}{\textbf{(S3) Compatible Curiosity}}: Tailored Curiosity-Driven Exploration for Value-Based RL}
\label{sec:curiosity_driven}
In previous works, the curiosity-driven exploration methods are always work with policy-based methods. In this section, we cast the key insight introduced above to RND~\citep{burda2018exploration} --- one of the leading algorithms for exploration --- to answer why is its vanilla design not suitable for value-based algorithms like Q-learning, and propose a variant of RND that is tailored for DQN.

The vanilla RND use two randomly initialized networks $\phi_{1}$ and $\phi_2$ to generate the intrinsic reward $r_{\mathrm{int}} = |\phi_{1}(s,a) - \phi_{2}(s,a)| \ge 0 $ for exploration. During learning, $\phi_2$ is a fixed network and the parameters of $\phi_1$ is optimized to approximate the output of $\phi_2$ for frequently-visited states. We follow \citet{burda2018exploration} to bound the intrinsic reward below $1$ and use $r_{\mathrm{int},t}$ to denote the intrinsic reward after $t$ step of optimization. Specifically, $r_{\mathrm{int},0}(s,a) = 1, \forall (s,a)$ --- an universally positive bonus is added to the primal reward function at beginning.

According to our analysis in previous sections, such a positive reward shift is equivalent to pessimistic initialization and will lead to conservative behaviors in $Q$-value estimation. Therefore, the exploration behaviors at the beginning of learning will be hindered, rather than boosted. 
To overcome such a conservative behavior induced by the pessimistic initialization, we proposed to use $r^-_{\mathrm{int}}(s,a) = |\phi_{1}(s,a) - \phi_{2}(s,a)|^2 - I, \forall s,a$, where $I$ is a positive constant that assures $r^-_{\mathrm{int}}\le 0$ is negatively initialized for optimistic exploratory behaviors.

\textbf{The Chicken-and-Egg Problem} To further understand the difference between the curiosity-driven method in ways of intrinsic reward and our reward shifting-based optimistic initialization, let us consider when do those curiosities work in each case: for intrinsic reward methods, an agent must \textbf{first experience} a new $(s,a)$ pair before receiving a high intrinsic reward --- this is extremely hard with an arg-max style policy. On the other hand, with optimistic initialization, the rarely-visited $(s,a)$ pairs will naturally have higher $Q$-values \textbf{before experiencing} it --- as the frequently-visited pairs have updated their values with a negatively shifted reward. From such a perspective, reward shifting not only works by itself motivates exploratory behaviors, but can also be seamlessly plugged into intrinsic reward methods to \textbf{encourage the first visitation} of new states.

\section{Experiments}
\begin{figure*}[t]
\centering
\begin{minipage}[htbp]{0.196\linewidth}
	\centering
	\includegraphics[width=1\linewidth]{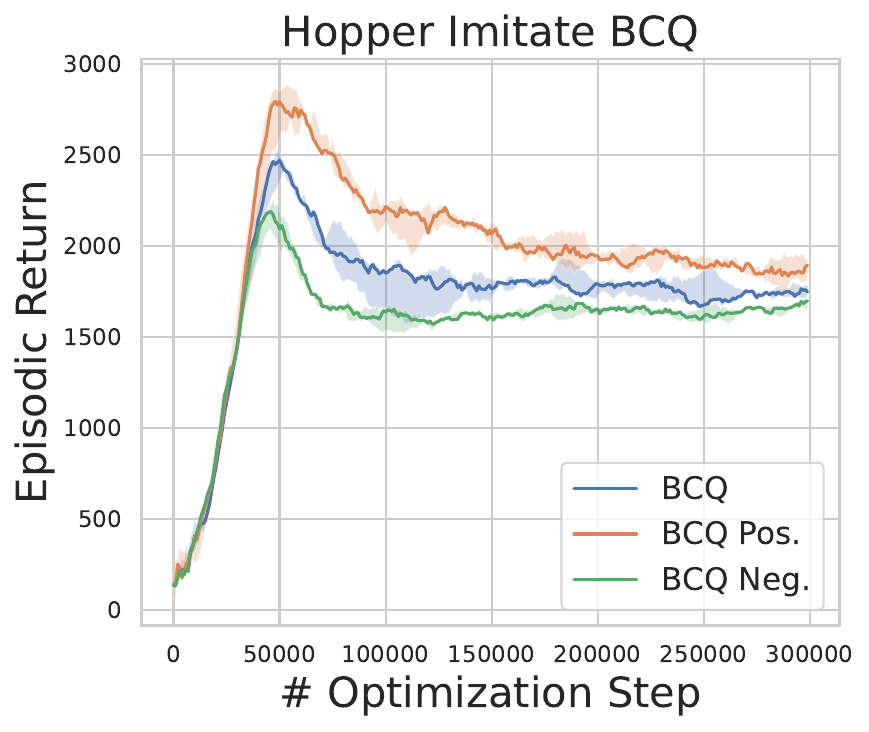}
\end{minipage}%
\begin{minipage}[htbp]{0.196\linewidth}
	\centering
	\includegraphics[width=1\linewidth]{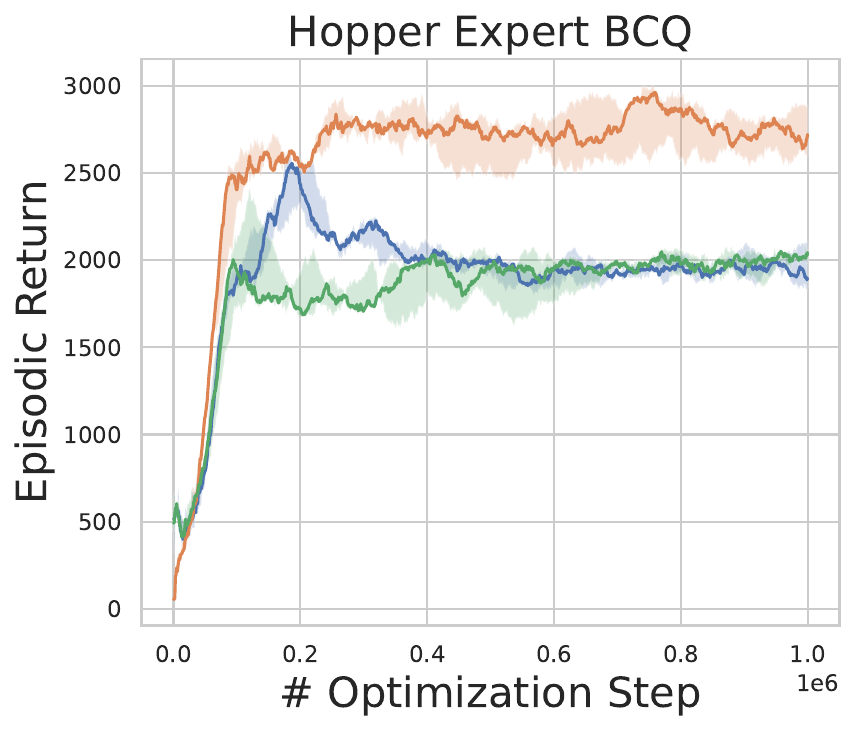}
\end{minipage}%
\begin{minipage}[left]{0.196\linewidth}
	\includegraphics[width=1\linewidth]{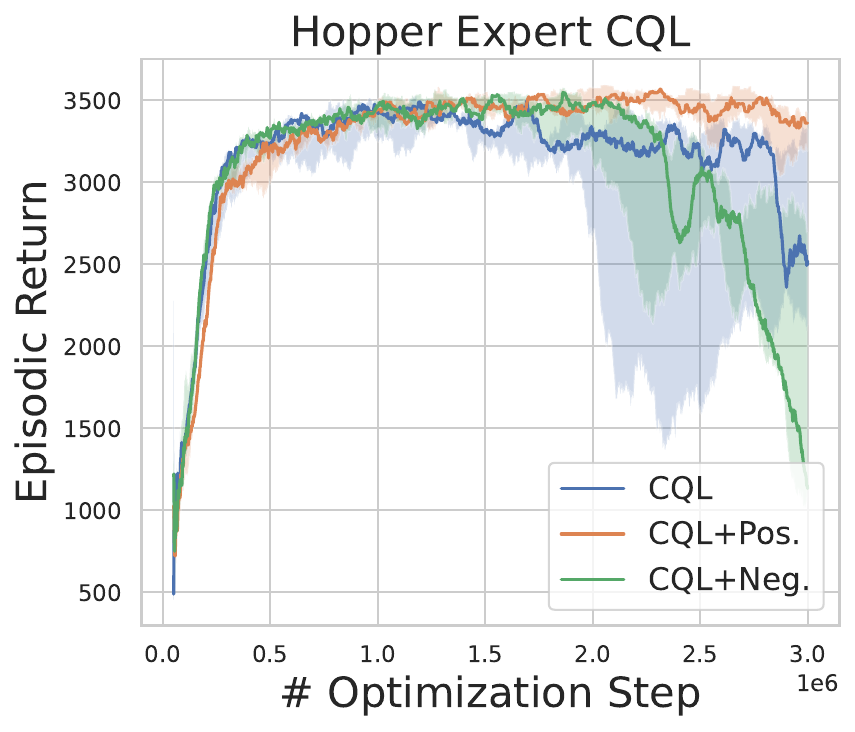}
\end{minipage}%
\begin{minipage}[htbp]{0.196\linewidth}
	\centering
	\includegraphics[width=1\linewidth]{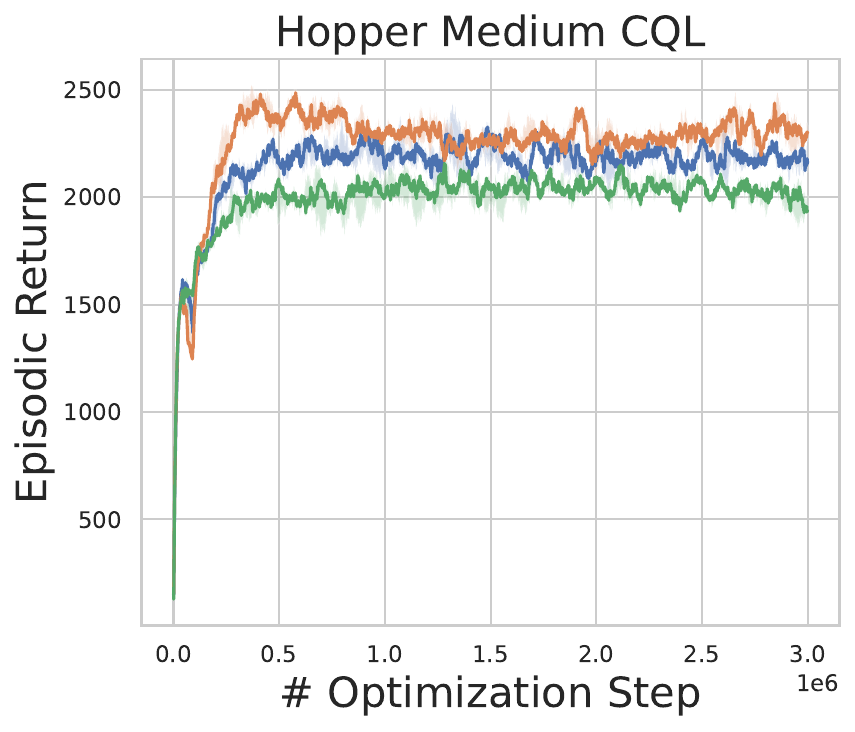}
\end{minipage}%
\begin{minipage}[left]{0.196\linewidth}
	\includegraphics[width=1\linewidth]{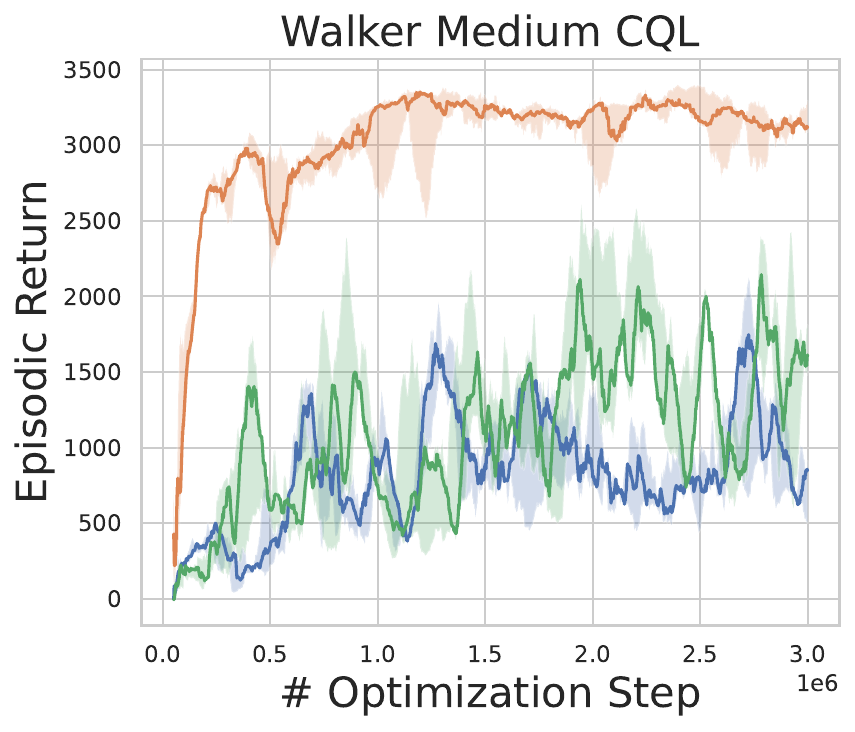}
\end{minipage}%
\caption{Results on offline RL settings. We verify our key insight that a positive reward shift equals to conservative exploitation thus helps offline value estimation, while a negative reward shift leads to worse performance. Results are from $10$ runs with shaded areas indicating the $25\%$-$75\%$ quantiles.}\label{fig:results_offline}
\end{figure*}

\subsection{\textcolor{purple}{\textbf{(S1)}}: Offline RL with Conservative $Q$-value Estimation}
\paragraph{Experiment Setup} 
We start our experiments by demonstrating the effectiveness of reward shifting in offline RL benchmarks. As discussed in Section~\ref{sec:offline}, a positive reward shift is equivalent to pessimistic initialization that benefits conservative exploitation. In general, our proposed method can be plugged in to any off-the-shelf offline RL algorithm, we choose to verify the effectiveness and generality of such a conservative $Q$-value estimation based on BCQ~\citep{fujimoto2018off} and CQL~\citep{bharadhwaj2020conservative}, i.e., both distribution-matching approach and conservative value estimation approaches in offline RL. To verify our insight, we experiment with both positive reward shift (\textbf{Pos.}) and negative reward shift (\textbf{Neg.}), added on either BCQ or CQL. 

\paragraph{Results}
Figure~\ref{fig:results_offline} shows our experiment results. We experiment with both the dataset generated in \citep{fujimoto2018off} (Hopper Imitate) and the dataset used in \citep{fu2020d4rl} (others), and find in our experiments that learning with the CQL dataset is much more stable. The first two panels show results with BCQ as the backbone algorithm. We observe a clear performance drop during the training of BCQ as the policy overfits the batched dataset. Differently, positive reward shift can alleviate such a problem and outperform BCQ in terms of both best-achieved performance and performance after convergence. The following three panels in Figure~\ref{fig:results_offline} use CQL as the backbone. 
Implementation details and more results can be found in Appendix~\ref{appdx:offline_detail}.

\paragraph{\textcolor{purple}{\textbf{Take-Away Message}}}
In all experiments, shifting the reward with a positive constant improves learning performance, while a negative reward shift impedes to efficient learning --- as expected. 

\begin{figure*}[t]
\centering
\begin{minipage}[htbp]{0.196\linewidth}
	\centering
	\includegraphics[width=1\linewidth]{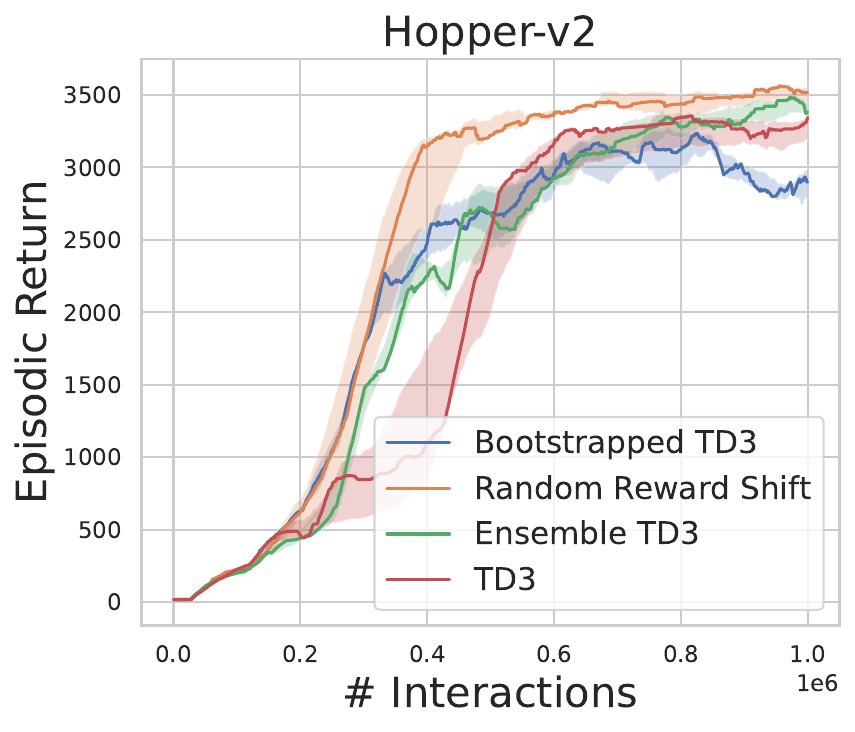}
\end{minipage}%
\begin{minipage}[htbp]{0.196\linewidth}
	\centering
	\includegraphics[width=1\linewidth]{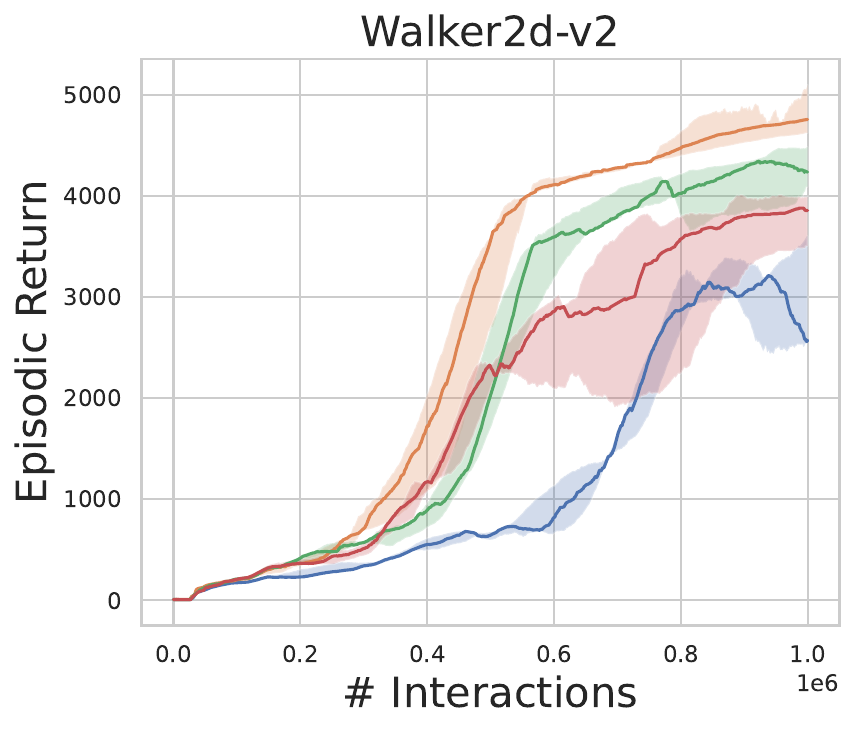}
\end{minipage}%
\begin{minipage}[htbp]{0.196\linewidth}
	\centering
	\includegraphics[width=1\linewidth]{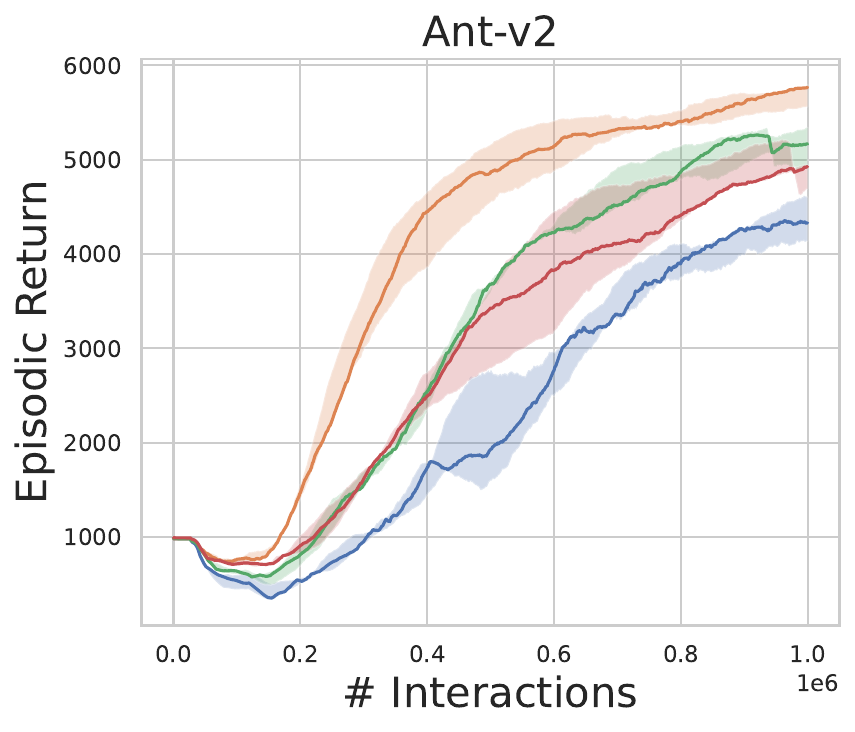}
\end{minipage}%
\begin{minipage}[h]{0.196\linewidth}
	\includegraphics[width=1\linewidth]{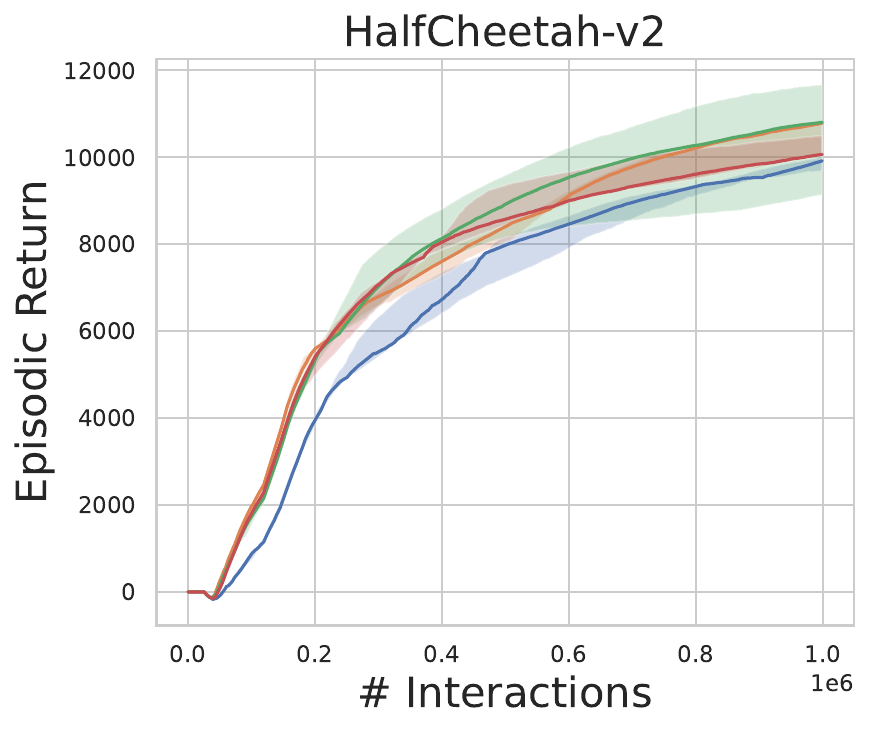}
\end{minipage}%
\begin{minipage}[h]{0.196\linewidth}
	\includegraphics[width=1\linewidth]{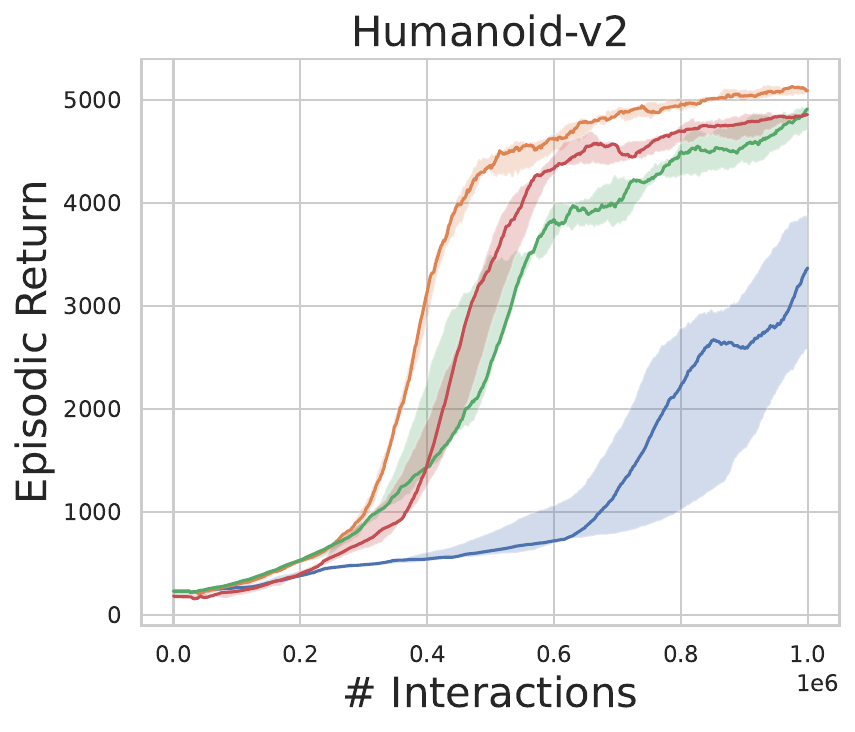}
\end{minipage}%
\caption{Results on continuous control tasks, Random Reward Shift (RRS) outperforms its value-based baselines in most environments. Results are from $10$ runs with shaded areas indicating the $25\%$-$75\%$ quantiles.}\label{fig:results_mujoco}
\vspace{-0.3cm}
\end{figure*}



\subsection{\textcolor{purple}{\textbf{(S2)}}: Online RL with Randomized Priors}
\paragraph{Experiment Setup}
We then conduct experiments in the MuJoCo locomotion benchmarks to demonstrate reward shifting improves learning efficiency in the online RL settings. As our implementation is based on TD3, we use TD3-based variants as our baselines: The \textbf{TD3} is trained with default settings according to ~\citet{fujimoto2018addressing}. We also include \textbf{Ensemble TD3} and \textbf{Bootstrapped TD3} as baselines due to they are similar to our work in using multiple $Q$-networks in value estimation. We follow ~\citet{osband2016deep} but extend it to the continuous control settings. In continuous control settings, the argmax operator is approximated by the policy network, and multiple policy networks are needed to cooperate with the multiple bootstrapped $Q$-value networks. Otherwise, multiple $Q$-value networks are not independent of each other thus breaking the condition of bootstrapped value estimation. The Ensemble TD3 presents the baseline performance when multiple $Q$-networks are used for value estimation in TD3, and works as an ablation of our method where all shift priors are set to be $0$. 

As has been illustrated in Sec.~\ref{sec:continuous_control}, learning with different reward shifting values is equivalent to learning with optimistic or conservative initialization. In our instantiating of RRS, we use $3$ $Q$-networks with different priors. We empirically find $\pm 0.5, 0$ work universally good for all environments. Though, future investigation on hyper-parameter may help to further improve the performance.
\paragraph{Results}
Results are shown in Figure~\ref{fig:results_mujoco}. RRS outperforms the vanilla TD3 in all five environments and outperforms all baseline methods in most tasks. In the experiment of Bootstrapped TD3 and Ensemble TD3, we also use $3$ $Q$-networks for a fair comparison. Note that there is a trade-off between computational complexity and sample efficiency, i.e., using more $Q$-networks may further improve the performance at the cost of more computational expenses, as reported in \citep{osband2016deep}. More implementation details, pseudo-code of RRS, and ablation studies can be found in Appendix~\ref{appdx:cc_detail}.

\paragraph{\textcolor{purple}{Take-Away Message}} Shifting the reward function can trade-off between exploration and exploitation. The ensemble performance of multiple value networks with random reward shifting drastically improve learning efficiency in continuous control.

\subsection{\textcolor{purple}{\textbf{(S3)}}: Optimistic Random Network Distillation}
\label{sec:curiosity_driven_exp}

\begin{figure*}[ht]
\centering
\begin{minipage}[htbp]{0.196\linewidth}
	\centering
	\includegraphics[width=1\textwidth]{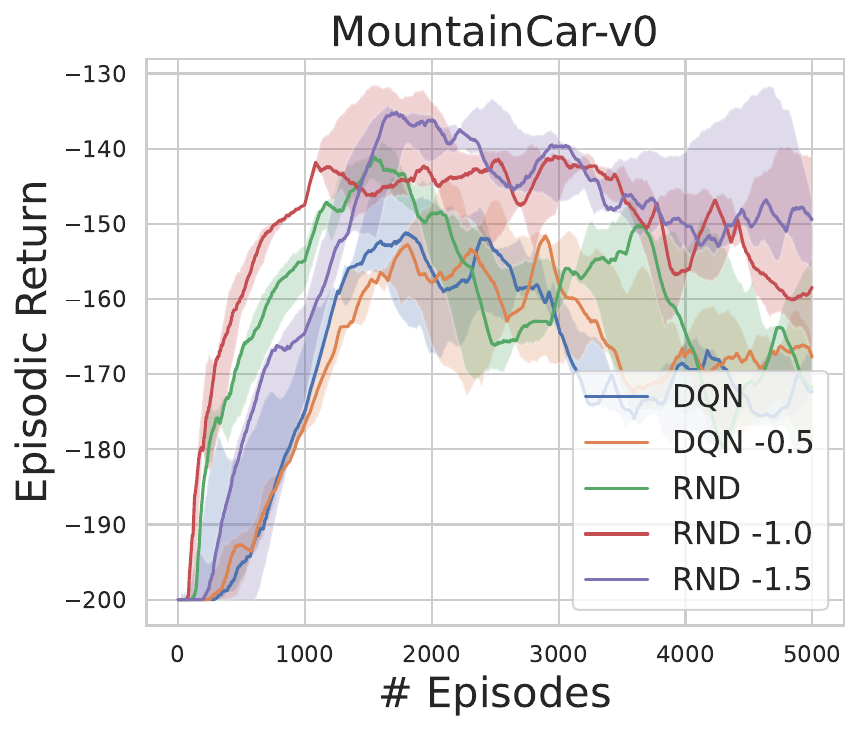}
\end{minipage}%
\begin{minipage}[htbp]{0.196\linewidth}
	\centering
	 \includegraphics[width=1\textwidth]{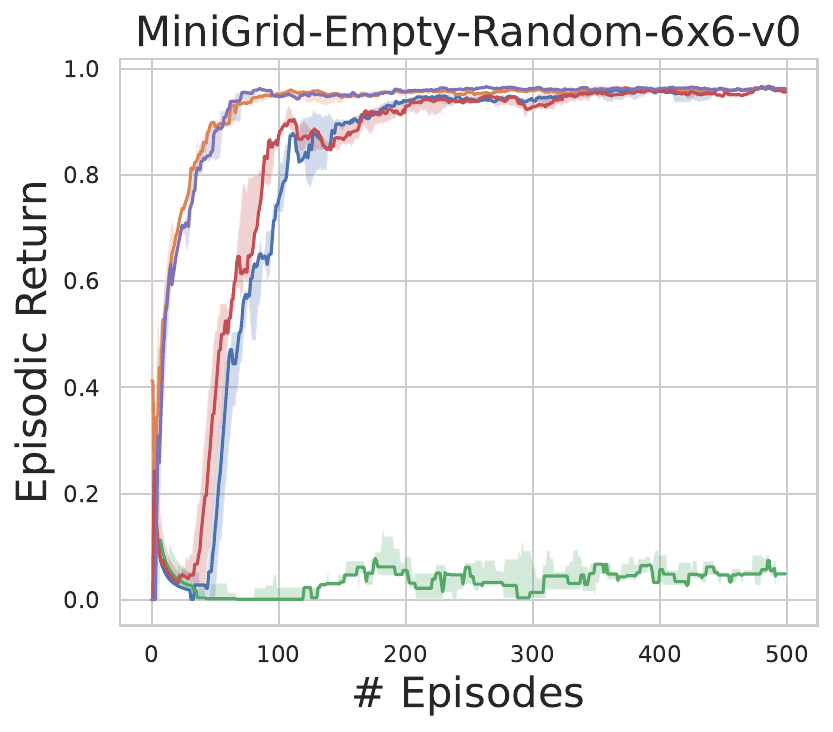}
\end{minipage}%
\begin{minipage}[htbp]{0.196\linewidth}
	\centering
	\includegraphics[width=1\textwidth]{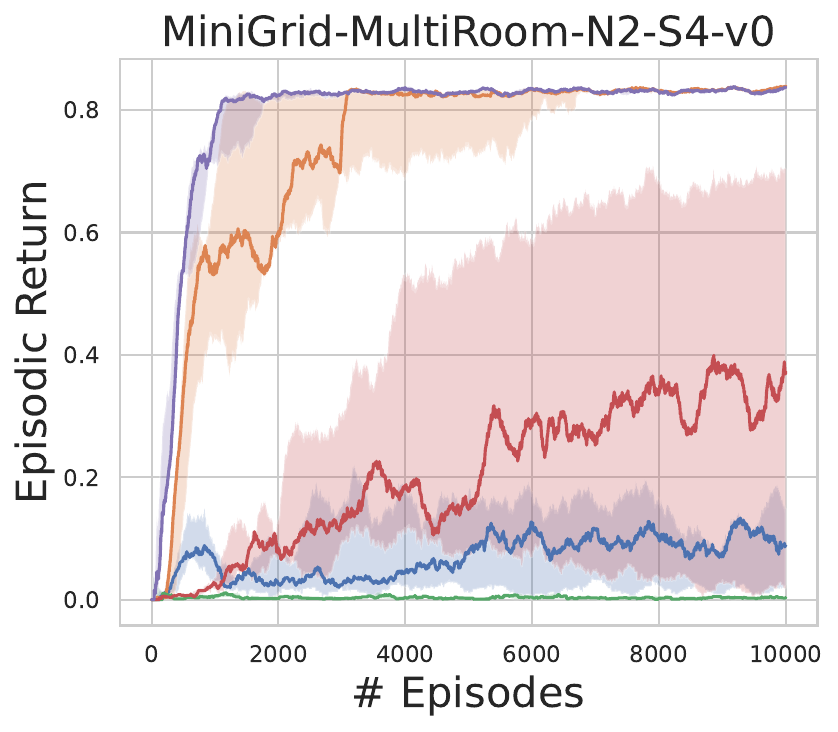}
\end{minipage}%
\begin{minipage}[h]{0.196\linewidth}
	\includegraphics[width=1\textwidth]{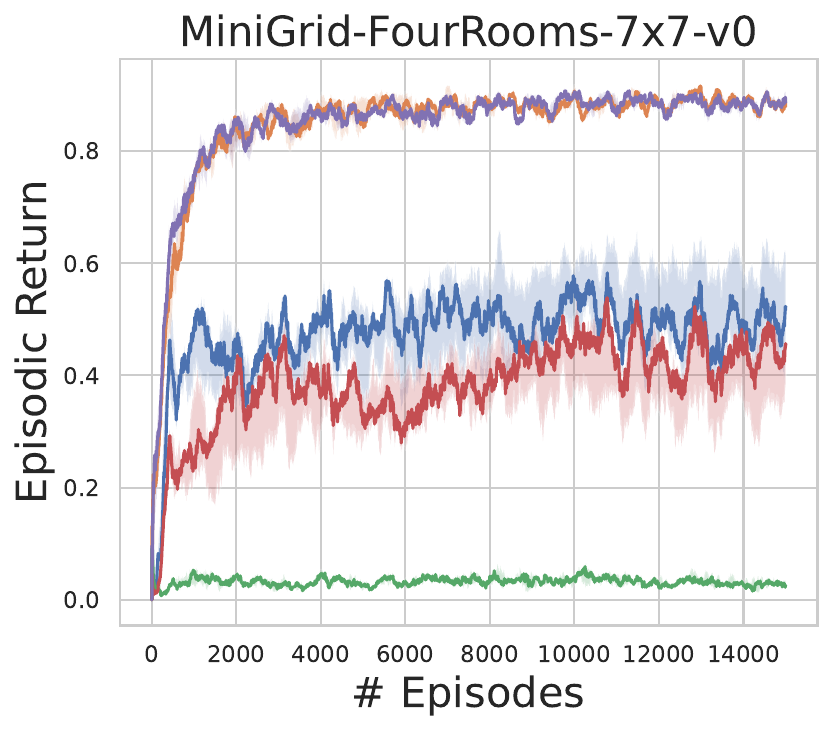}
\end{minipage}%
\begin{minipage}[h]{0.196\linewidth}
	\includegraphics[width=1\textwidth]{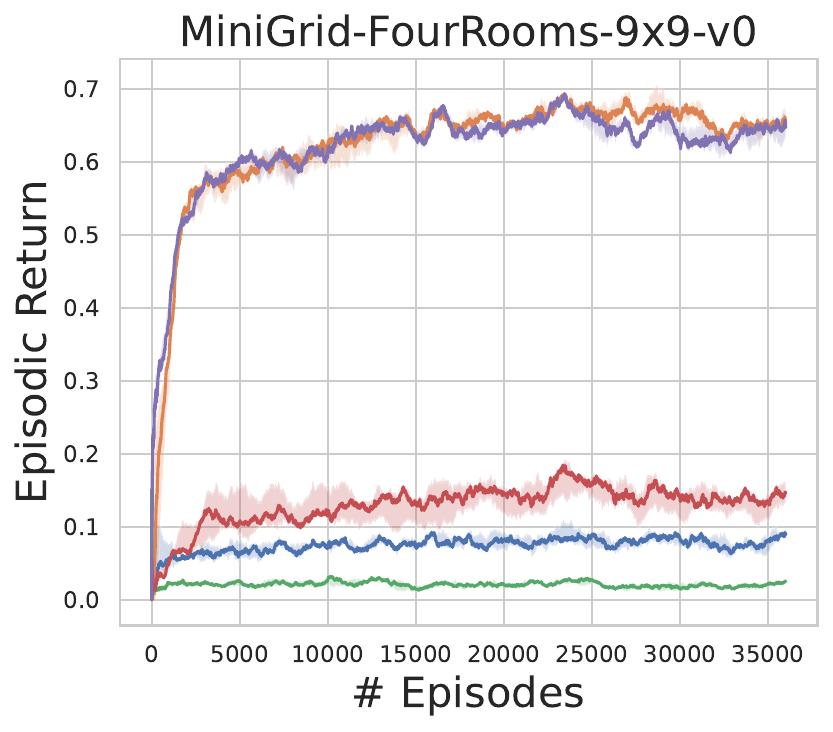}
\end{minipage}%
\caption{Value-based RND with shifted prior: Plugging the vanilla RND into DQN is not well-motivated according to our analysis in Section~\ref{sec:curiosity_driven}. The insight of equivalence between negative reward shifting and curiosity-driven exploration motivates us to negatively shift the intrinsic reward of RND, which drastically improves DQN-based RND. Results are from $10$ runs with shaded areas indicating the $25\%$-$75\%$ quantiles.}\label{fig:results_RND_new}
\end{figure*}
\paragraph{Experiment Setting}
To demonstrate the key insight and effectiveness of reward shifting in optimistic exploration, we benchmark with five discrete exploration tasks, including the classic MountainCar control and four MiniGrid navigation tasks~\citep{gym_minigrid}, environment details can be found at Appendix~\ref{appdx:RND}.

\paragraph{Results}
Figure~\ref{fig:results_RND_new} presents the results of $5$ different methods for comparison: besides the as-is \textcircled{\raisebox{-0.9pt}{1}}\textbf{DQN} and \textcircled{\raisebox{-0.9pt}{2}}\textbf{RND} baselines, we use \textcircled{\raisebox{-0.9pt}{3}}\textbf{DQN -0.5} to indicate DQN with a $-0.5$ reward shift, and \textcircled{\raisebox{-0.9pt}{4}}\textbf{RND -1.0} \textcircled{\raisebox{-0.9pt}{5}}\textbf{ RND -1.5} to indicate RND with $-1.0, -1.5$ reward shift, respectively. In the following, we use $\succ$ between methods to indicate the former outperforms the latter. Comparing the results:

\textbf{1. reward shifting is equivalent to optimistic initialization, thus boosting exploration} \\
\textcircled{\raisebox{-0.9pt}{3}} $\succ$ \textcircled{\raisebox{-0.9pt}{1}}: a negative reward shift is equivalent to optimistic initialization and helps exploration.

\textbf{2. RND is effective for value-based exploration --- as long as a negative intrinsic reward is used}
\textcircled{\raisebox{-0.9pt}{1}} $\succ$ \textcircled{\raisebox{-0.9pt}{2}}: vanilla RND with positive intrinsic reward is always worse than DQN: adding a positive intrinsic reward like RND is harmful for exploration as it is equivalent to a pessimistic initialization. \\
\textcircled{\raisebox{-0.9pt}{4}} $\succ$\textcircled{\raisebox{-0.9pt}{1}}; \textcircled{\raisebox{-0.9pt}{5}} $\succ$ \textcircled{\raisebox{-0.9pt}{3}}: RND is effective for exploration (i.e., improve over DQN) as long as the intrinsic reward bonus is always negative.

\textbf{3. reward shifting for optimistic initialization is orthogonal to other exploration methods}\\
\textcircled{\raisebox{-0.9pt}{5}} $\succ$ \textcircled{\raisebox{-0.9pt}{4}}: increasing the magnitude of the negatively shifted reward can further improve the exploration performance — reward shifting can either work in isolation or get combined with other exploration algorithms as they are working orthogonally.~\footnote{Code is available at \href{https://github.com/holarissun/RewardShifting}{GitHub}}

\paragraph{\textcolor{purple}{Take-Away Message}}
Reward shifting is equivalent to optimistic initialization, it can help with exploration in value-based methods. Importantly, such an intrinsic motivation is orthogonal to previous count-based methods, such that it can either work in isolation or be combined with conventional curiosity-driven exploration methods.



\section{Conclusion}
In this work, we studied how reward shifting affects policy learning in value-based deep reinforcement learning algorithms. Although constant reward shift should not change the optimal policy induced by the optimal value function, in practice such a constant shift \textit{does affect the function approximation}, and leads to different learning behaviors. Our detailed analysis manifests the fact that a constant reward shift is equivalent to using different initialization in the value function approximation. The proposed idea is then verified through a variety of application settings.
Specifically, we show that a negative reward shift leads to curiosity-driven exploration, while a positive reward shift helps conservative exploitation. Importantly, our analysis reveals that changing reward shifting constant itself is sufficient in trading-off between exploration and exploitation. We empirically verify the effectiveness of the proposed method in a variety of experiments, including better exploitation in offline RL, sample-efficient learning in continuous control benchmarks, and enhanced curiosity-driven exploration in value-based discrete control.

While our experiments demonstrate the performance gain is quite robust to the shifting constant, we would like to point out that the theoretical guidance for such a shifting constant is missing. Potential solutions may lie in analysis from the perspective of optimization for black-box models, yet it is out of the scope of the current empirical study and left for future research.

\section{Acknowledgement}
We thank all anonymous reviewers, ACs, PCs for their efforts and time in the reviewing process and in improving our paper. This work is done with the warm supports from the MMLab members. We acknowledge the insightful discussions with Ziping Xu, the Hot Spring Harbor group and the van der Schaar Lab members in improving the presentations of this paper. We thank Takuya Kanazawa in pointing out the concurrent work of \citet{dubey2022pursuit} that also discusses the effects of shifting terms in reward function.



\newpage
\bibliography{iclr2022_conference}
\bibliographystyle{unsrtnat}

\medskip

\section*{Checklist}


\begin{enumerate}

\item For all authors...
\begin{enumerate}
  \item Do the main claims made in the abstract and introduction accurately reflect the paper's contributions and scope?
    \answerYes{}
  \item Did you describe the limitations of your work?
    \answerYes{We discussed in the Conclusion that this work is mainly limited in empirical studies.}
  \item Did you discuss any potential negative societal impacts of your work?
    \answerNA{Our research studies the general problem of reward shifting in reinforcement learning, aiming at improve learning efficiency.}
  \item Have you read the ethics review guidelines and ensured that your paper conforms to them?
    \answerYes{}
\end{enumerate}

\item If you are including theoretical results...
\begin{enumerate}
  \item Did you state the full set of assumptions of all theoretical results?
    \answerYes{}
        \item Did you include complete proofs of all theoretical results?
    \answerYes{Please refer to Appendix~\ref{appdx:proof}}
\end{enumerate}

\item If you ran experiments...
\begin{enumerate}
  \item Did you include the code, data, and instructions needed to reproduce the main experimental results (either in the supplemental material or as a URL)?
    \answerYes{Implementation details can be found at Appendix~\ref{appdx:implementation_details}. Code is open-sourced at \href{https://github.com/holarissun/RewardShifting}{https://github.com/holarissun/RewardShifting}}
  \item Did you specify all the training details (e.g., data splits, hyperparameters, how they were chosen)?
    \answerYes{Implementation details can be found at Appendix~\ref{appdx:offline_detail}, \ref{appdx:cc_detail}, and \ref{appdx:RND}}
        \item Did you report error bars (e.g., with respect to the random seed after running experiments multiple times)?
    \answerYes{In our experiments, the results are averaged over 10 runs, and shaded areas indicate the $25\%$ - $75\%$ quantile values.}
        \item Did you include the total amount of compute and the type of resources used (e.g., type of GPUs, internal cluster, or cloud provider)?
    \answerYes{Please refer to Appendix~\ref{appdx:implementation_details}.}
\end{enumerate}

\item If you are using existing assets (e.g., code, data, models) or curating/releasing new assets...
\begin{enumerate}
  \item If your work uses existing assets, did you cite the creators?
    \answerYes{\cite{fujimoto2018off, bharadhwaj2020conservative}}
  \item Did you mention the license of the assets?
    \answerYes{MIT license.}
  \item Did you include any new assets either in the supplemental material or as a URL?
    \answerNo{}
  \item Did you discuss whether and how consent was obtained from people whose data you're using/curating?
    \answerYes{We follow BCQ and CQL to use the dataset they used in the papers.}
  \item Did you discuss whether the data you are using/curating contains personally identifiable information or offensive content?
    \answerNA{}
\end{enumerate}

\item If you used crowdsourcing or conducted research with human subjects...
\begin{enumerate}
  \item Did you include the full text of instructions given to participants and screenshots, if applicable?
    \answerNA{Not applicable.}
  \item Did you describe any potential participant risks, with links to Institutional Review Board (IRB) approvals, if applicable?
    \answerNA{Not applicable.}
  \item Did you include the estimated hourly wage paid to participants and the total amount spent on participant compensation?
    \answerNA{Not applicable.}
\end{enumerate}
\end{enumerate}


\appendix
\onecolumn
\section{Extended Related Work}
We extend our related work section on the following related topics as suggested by reviewers:
\subsection{Discussion on Ensembles and Distributional RL}
In Distributional RL literature~\citep{bellemare2017distributional, dabney2018distributional, lyle2019comparative, barth2018distributed, dabney2020distributional}, the distribution of the $Q$-value, rather than the mean scaler, is estimated. Distributional-RL focuses on stochastic reward mechanisms and smooth the temporal difference learning in a distributional level, and can be applied to risk-sensitive scenarios~\citep{urpi2021risk} where the worst-case performance can be controlled~\citep{deletang2021model}. In those scenarios, the systematic uncertainty is the crucial issue to address, whereas in our work, we focus on deterministic transition dynamics and use OFU to tackle the epistemic uncertainty in the section of RRS.

Several previous works discussed ensemble methods for exploration, both for discrete control~\citep{chen2017ucb,lee2021sunrise} and for continuous control~\citep{an2021uncertainty, chen2021randomized}. However, in our work we show that more exploratory behavior can emerge with the help of reward shifting under only single $Q$-value network — as a proof of concept that negative reward shift is equivalent to optimistic initialization.

\subsection{Model-Based Exploration and Uncertainty Estimation}
Besides the model-free value-based RL methods we focused in our paper, there exist literature like R-Max~\citep{brafman2002r} that work with model-based methods for better exploration. Moreover, combining multiple neural networks for uncertainty estimation is well-established in supervised learning~\citep{lakshminarayanan2017simple}, and similari idea has been explored in the context of RL for discrete control~\citep{osband2016deep,sun2022daux}.
In this work, we showcase that a diversified set of reward shifting constants can work as priors for such an ensemble.

Given the ensemble networks without ground-truth, it is in general hard to disentangle the aleatoric uncertainty from epistemic uncertainty. Our method works on deterministic reward and transition dynamics to circumvent the discussion of uncertainty stratification. In the deterministic settings, the source of uncertainty can be solely attributed to the epistemic uncertainty and hence help informative exploration. When it comes to the stochastic environments, the entanglement of aleatoric uncertainty and epistemic uncertainty will make the problem much more difficult~\citep{clements2019estimating} as intrinsic motivation methods may get trapped by pursuing the actions that result in high aleatoric uncertainty in certain circumstances (e.g., the Noisy-TV)~\citep{burda2018large}.

\section{Proof of Proposition~\ref{prop1}}
\label{appdx:proof}
\begin{proof}

the estimated $Q$-value $\hat{Q}(s,a)$ is composed by the two estimators with function approximation error, defined as $\epsilon_{b^+}(s,a) = \tilde{Q}_{b^+,t}(s,a) - \frac{b^+}{1-\gamma} - Q^*(s,a)$, and $\epsilon_{b^-}(s,a) =  \tilde{Q}_{b^-,t}(s,a) - \frac{b^-}{1-\gamma} - Q^*(s,a)$. 

\begin{equation}
\begin{split}
    &\quad  (1-\beta ) \epsilon_{A,t} + \beta \epsilon_{B,t} \\ 
    & = 2\eta(1-\beta) \hat{Q}^*  + (1-\beta)(1-2\eta)\tilde{Q}_{A,t} + 2\eta\beta \hat{Q}^*  + \beta(1-2\eta)\tilde{Q}_{B,t} \\
    & = 2\eta \hat{Q}^* + (1-2\eta) [(1-\beta)\tilde{Q}_{A,t} + \beta \tilde{Q}_{B,t}]  \\
    & = 2\eta \hat{Q}^* + (1-2\eta) [(1-\beta) (1-2\eta)^t \tilde{Q}_{A,0} + \frac{4(1-\beta)\eta^2}{1-(1-2\eta)^t}\hat{Q}^* + \beta (1-2\eta)^t \tilde{Q}_{B,0} + \frac{4\beta\eta^2}{1-(1-2\eta)^t}\hat{Q}^*] \\
    & = 2\eta \hat{Q}^* + (1-2\eta)[(1-2\eta)^t((1-\beta)\tilde{Q}_{A,0} + \beta\tilde{Q}_{B,0}) + \frac{4\eta^2}{1-(1-2\eta)^t}\hat{Q}^*] \\
    & = \epsilon_{C, t}
\end{split}
\end{equation}
where $C = (1-\beta)A + \beta B$ and the last line requires $\tilde{Q}_{A,0}=\tilde{Q}_{B,0}=\tilde{Q}_{C,0}$ are identical initialization.

With this notion, Equation (\ref{eqn:ofu4}) can be re-written as 
\begin{equation}
\begin{split}
    \hat{Q}(s,a) & = Q^*(s,a) + (1-\beta) \epsilon_{b^+}(s,a) + \beta \epsilon_{b^-}(s,a) \\
    & = Q^*(s,a) + \epsilon_{(1-\beta)b^+ + \beta b^-}(s,a)  \\
    & = Q^*(s,a) + \epsilon_{c_r}(s,a)  
\end{split}
\end{equation}

where the second line relies on the linear assumption of the approximation error $(1-\beta) \epsilon_{b^+}(s,a) + \beta \epsilon_{b^-}(s,a)$. We further have $(1-\beta)\tilde{Q}_{b^+} + \beta\tilde{Q}_{b^-} = \tilde{Q}_{(1-\beta)b^+ + \beta b^-}$ and $\hat{Q}(s,a) = \tilde{Q}_{c_r}(s,a)$, telling us that trading-off between the constant $b^-$ used for exploration and the constant $b^+$ used for exploitation with the coefficient $\beta$ is equivalent to use another constant with value of $c_r = (1-\beta)b^+ + \beta b^-$.

\end{proof}

\section{Implications of Assumption in Section~\ref{sec:continuous_control}}
\label{appdx:implication}

In our main text, the estimated values for extremely o.o.d. samples are assumed to be near zeros. We provide detailed implications and explanations in this section.

On the one hand, it’s clear that such an assumption holds for the tabular settings, that un-visited state-action pairs have the value in tabular initialization.

On the other hand, we acknowledge it as a mild assumption that there always exists o.o.d. samples that have the $Q$-values near zero for function approximation settings. Interpolation between those o.o.d. samples and other state-action pairs will clearly lead to an ``in-between’’ value estimation, which in practice can be achieved with properly regularized neural networks.

The key insight we want to emphasize in Section~\ref{sec:continuous_control} is that for frequently visited state-action pairs, the value discrepancy with different initialization are small, while for seldomly-visited state-action pairs, the discrepancy are relatively large, enabling the usage of such discrepancy as exploration bonus.

\section{Implementation Details and Ablation Studies}
\label{appdx:implementation_details}
\paragraph{Hardware and Training Time}
We experiment on a server with 8 TITAN X GPUs and 32 Intel(R) E5-2640 CPUs. In general, shifting the reward does not introduce further computation burden except in the continuous control tasks, our method of Random Reward Shift (RRS) requires two additional $Q$-value networks. In our PyTorch-based implementation, those additional networks can be easily implemented and optimized in a parallel manner, and the extra computational burden is equivalent to using a $\sqrt{3}$ times wider neural network during optimization. It is worth noting that RRS is computationally much cheaper than the Bootstrapped TD3, where additional policy networks are also needed.

\paragraph{Network Structure}
Our implementation of TD3, BCQ and CQL are based on code released by the authors, without changing hyper-parameters. We implement DQN based on a 3-layer fully connected neural network with $64$ hidden units for the $Q$-value function, using ReLU and linear activation respectively. We use the Adam optimizer with learning rate of $0.001$, and use an epsilon-greedy approach as naive exploration strategy. In our RND, we use two 4-layer fully connected neural networks with $512$ units and ReLU activation in each hidden layer, and a softmax activation for the output layer. Adam optimizer is used for the optimization of the RND networks with learning rate $0.0001$.

Our code is provided in the supplementary materials, and will be made public available.

\subsection{Offline RL}
\label{appdx:offline_detail}

In our experiments, we use a fixed dataset with $10$k offline trainsition tuples for offline RL learning. Our implementation of BCQ and CQL are both based on the code provided by the authors. The only change we made to verify our insight is to shift the reward by a constant. In most environments, we find $r'=r+8$ provides good enough performance. While in Hopper Medium CQL we find using a smaller positive reward shift $r'=r+1$ works better than $r'=r+8$, and for Walker Medium CQL, using a larger reward shift of $r'=r+50$ further improves the result with $r'=r+8$. 

\begin{figure}[h]
\centering
\begin{minipage}[htbp]{0.33\linewidth}
	\centering
	\includegraphics[width=1\linewidth]{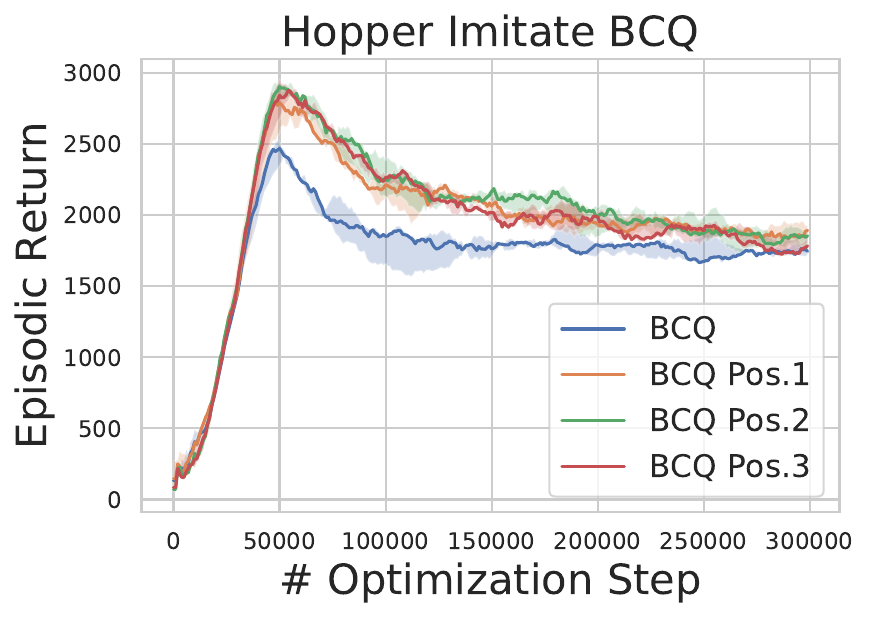}
\end{minipage}%
\begin{minipage}[htbp]{0.33\linewidth}
	\centering
	\includegraphics[width=1\linewidth]{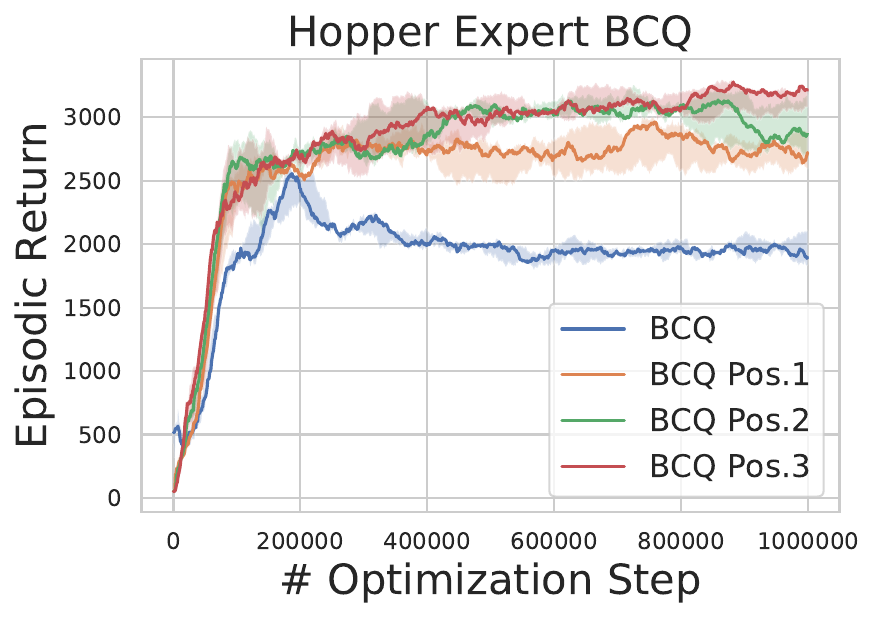}
\end{minipage}%
\begin{minipage}[htbp]{0.33\linewidth}
	\centering
	\includegraphics[width=1\linewidth]{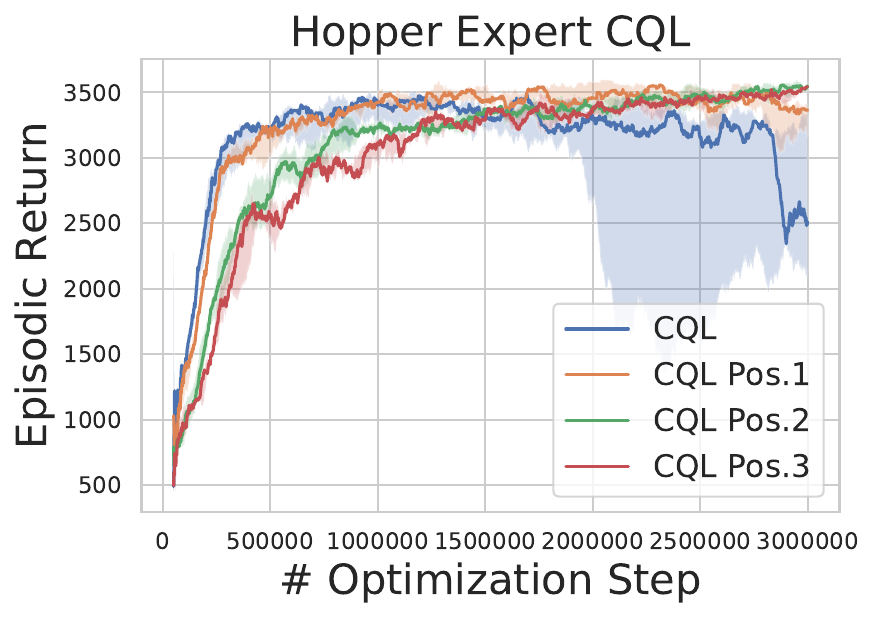}
\end{minipage}\\%
\begin{minipage}[left]{0.33\linewidth}
	\includegraphics[width=1\linewidth]{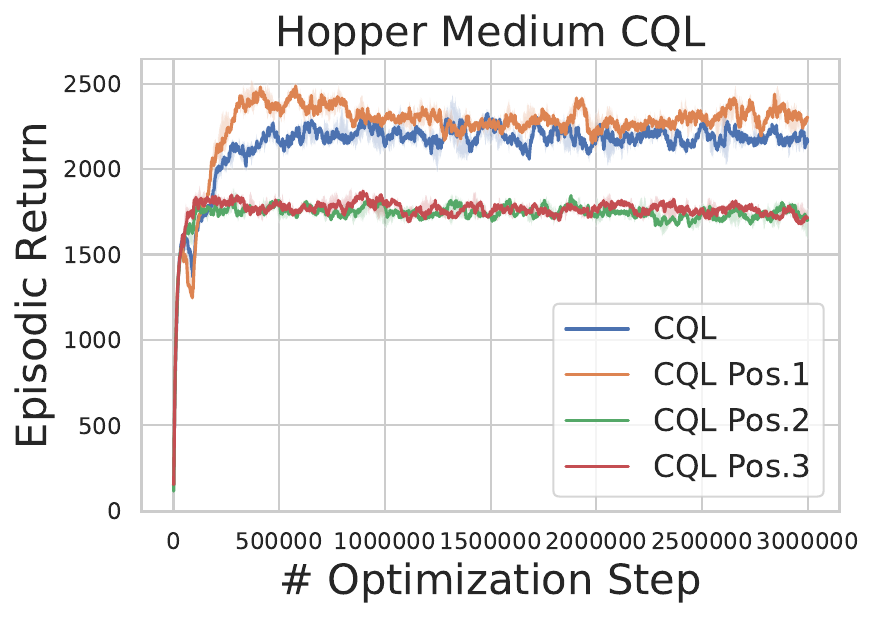}
\end{minipage}%
\begin{minipage}[left]{0.33\linewidth}
	\includegraphics[width=1\linewidth]{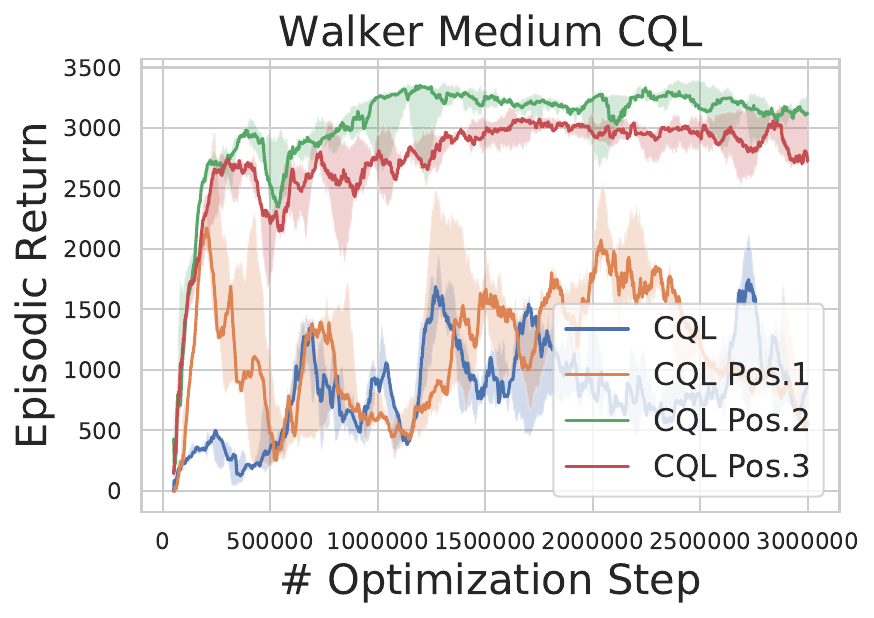}
\end{minipage}%
\caption{Performance with different reward shift constants.}
\label{fig:ablation_offline}
\end{figure}

Figure~\ref{fig:ablation_offline} shows different performance under different choices of the reward shift constant. We denote a positive reward shift $r' = r + 8$ as \textbf{Pos.1}, denote $r'=r+20$ as \textbf{Pos.2} and denote $r'=r+50$ as \textbf{Pos.3} for all experiments excetp in the Hopper Medium CQL we use \textbf{Pos.1} to denote $r'=r+1$. 

In the experiments based on BCQ (first two figures). We can observe a uniformly performance improvement with all choices of reward shift constants. As the algorithm of CQL has already taken the conservative value estimation into consideration, in the experiments based on CQL, the performance is more closely related to the constant we use. Specifically, in Hopper Expert, while using any of the positive reward shift constants improve the learning stability, $r'=r+8$ performs better on preserving the learning efficiency during early learning stage. For Hopper Medium, we find using larger positive constants hinder the performance. For Walker Medium, using a larger constant in reward shift performs much better than using a smaller one.

\subsection{Continuous Control}
\label{appdx:cc_detail}


\paragraph{Details of RRS}

\begin{algorithm}[t]
\fontsize{8.5}{8.5}\selectfont
	\caption{Sample-Efficient Continuous Control with Random Reward Shift}
	\label{algo_RRS}
	\begin{algorithmic}
		\STATE \textbf{Require}
		 \STATE ~~~ Size of mini-batch $N$, smoothing factor $\tau > 0$, $K$ reward shift values $r'_{k} = r + b_k, k=1, \dots,K$.
		\STATE ~~~ Random initialized policy network $\pi_{\theta}$, target policy network $\pi_{\theta'}$, $\theta'\leftarrow  \theta$.
		 \STATE ~~~$K$ random initialized $Q$ networks, and corresponding target networks, parameterized by $w_{k},w'_{k}$, $w'_k\leftarrow w_{k}$ for $ k = 1, \dots, K$. (e.g., a ModuleList in PyTorch).
		\FOR{iteration $= 1,2,...$}
		\STATE Uniformly sample one of the $K$ $Q$-functions, $Q_{w_k}$, for policy update
		\FOR{t $= 1,2,...$}
		\STATE $\#$ Interaction
		\STATE Run policy $\pi_{\theta}$, and collect transition tuples $(s_t,a_t,s'_t,r_t)$.
		\STATE Sample a mini-batch of transition tuples $\{(s,a,s',r)_{i}\}_{i = 1}^N$.
		\STATE $\#$ Update $Q_w$ (in parallel)
	    \STATE Calculate the $k$-th target $Q$ value $y_{k,i} = r_i + b_k +  Q_{w'_{k}}(s'_i,\pi_{\theta'}(s'_i))$
	    \STATE Update $w_{k}$ with loss $\sum_{i = 1}^N (y_{k,i} - Q_{w_{k}}(s_i,a_i))^2$.
		\STATE $\#$ Update $\pi_\theta$
	    \STATE Update policy $\pi_\theta$ with $Q_{w_k}$
		\ENDFOR
		\STATE $\#$ Update target networks
		\STATE $\theta'\leftarrow \tau \theta + (1-\tau) \theta'$.
		\STATE $w_{k}'\leftarrow \tau w_{k} + (1-\tau) w_{k}', k=1,\dots,K$.
		\ENDFOR
	\end{algorithmic}
\end{algorithm}

Although we find in the motivating example that a $-5$ reward shift is able to remarkably improve the asymptotic performance of TD3, in this work we aim at proposing an uniformly suitable method based on the insight behind the motivating example. Therefore we propose to use $\pm 0.5, 0$ as the reward shifting constants. We find in experiment that the sampling frequency does not affect the performance. And in the experiments we follow BDQN~\cite{osband2016deep} to use a fixed value network throughout a whole trajectory. i.e., one of the $K$ $Q$-networks is sampled uniformly after each episode with length of $1000$ timesteps. Intuitively, searching for more suitable reward randomization designs may further improve the performance, yet that is beyond the coverage of this work. 

\begin{figure}[h]
\centering
\begin{minipage}[htbp]{0.33\linewidth}
	\centering
	\includegraphics[width=1\linewidth]{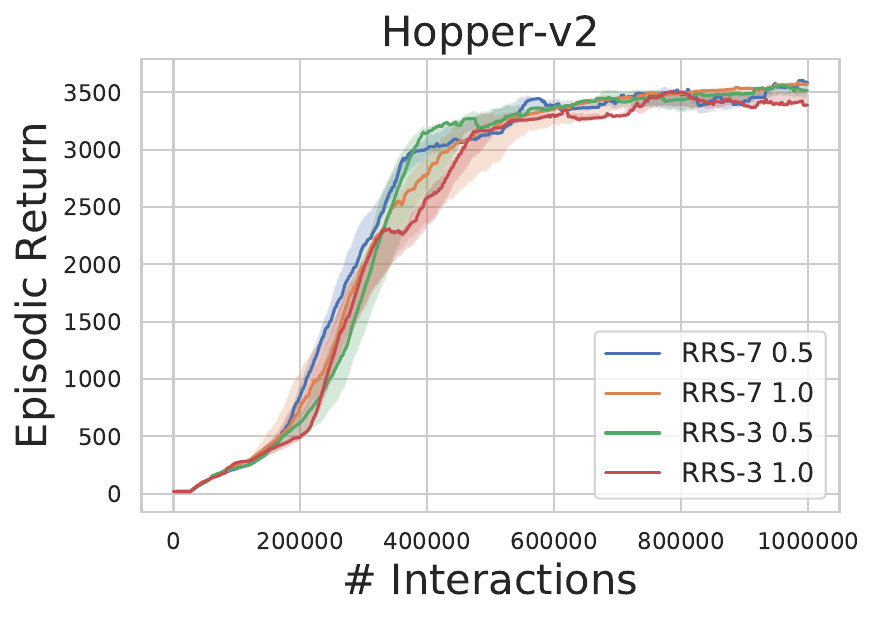}
\end{minipage}%
\begin{minipage}[htbp]{0.33\linewidth}
	\centering
	\includegraphics[width=1\linewidth]{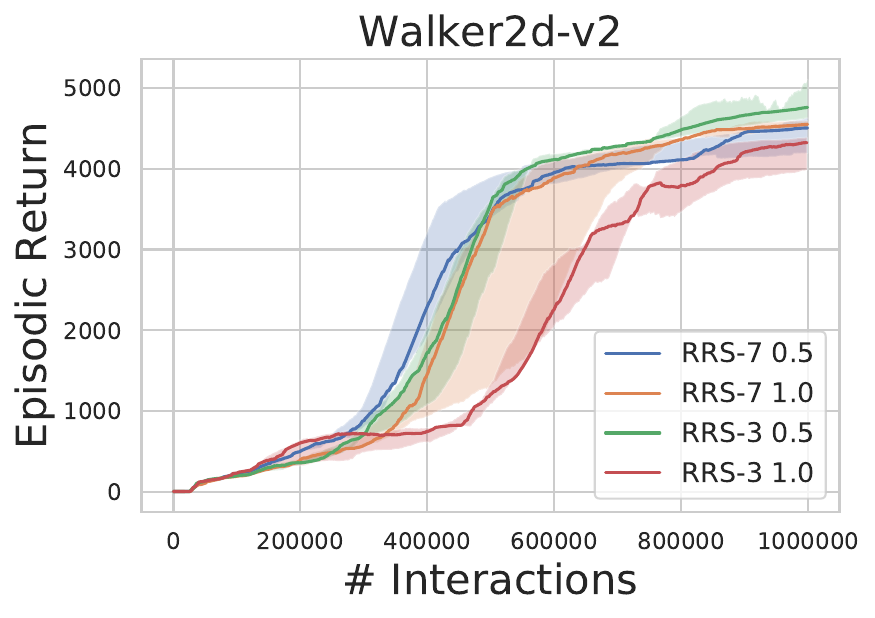}
\end{minipage}%
\begin{minipage}[htbp]{0.33\linewidth}
	\centering
	\includegraphics[width=1\linewidth]{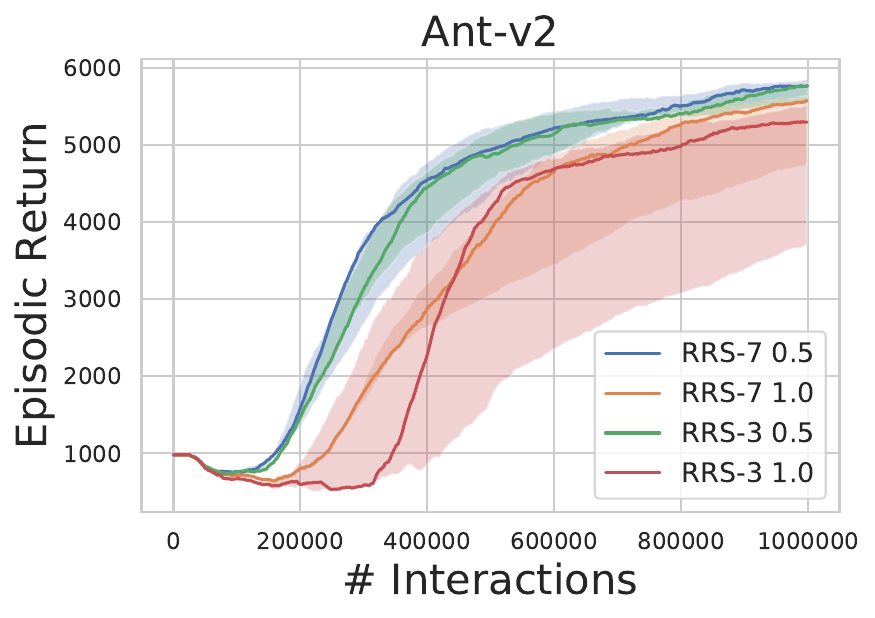}
\end{minipage}\\%
\begin{minipage}[left]{0.33\linewidth}
	\includegraphics[width=1\linewidth]{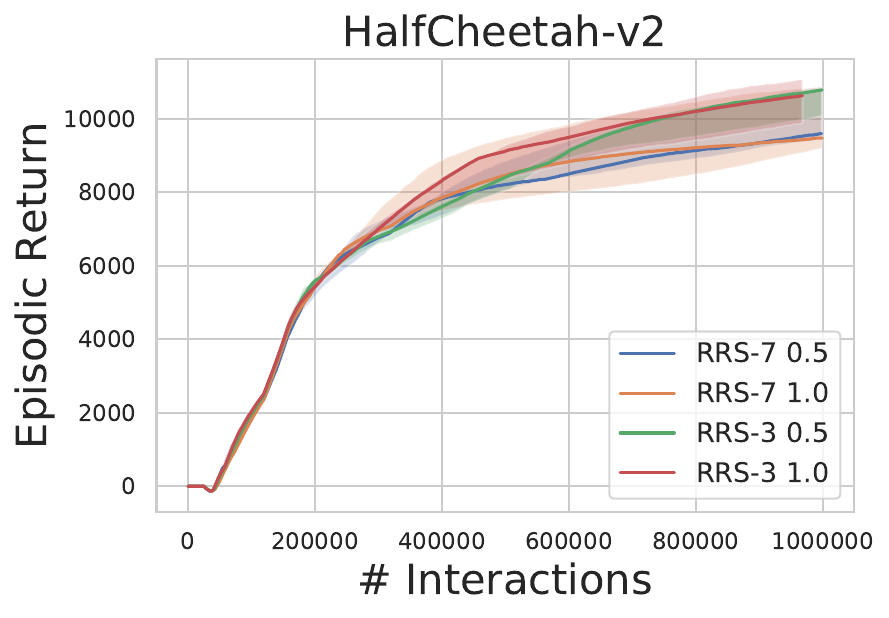}
\end{minipage}%
\begin{minipage}[left]{0.33\linewidth}
	\includegraphics[width=1\linewidth]{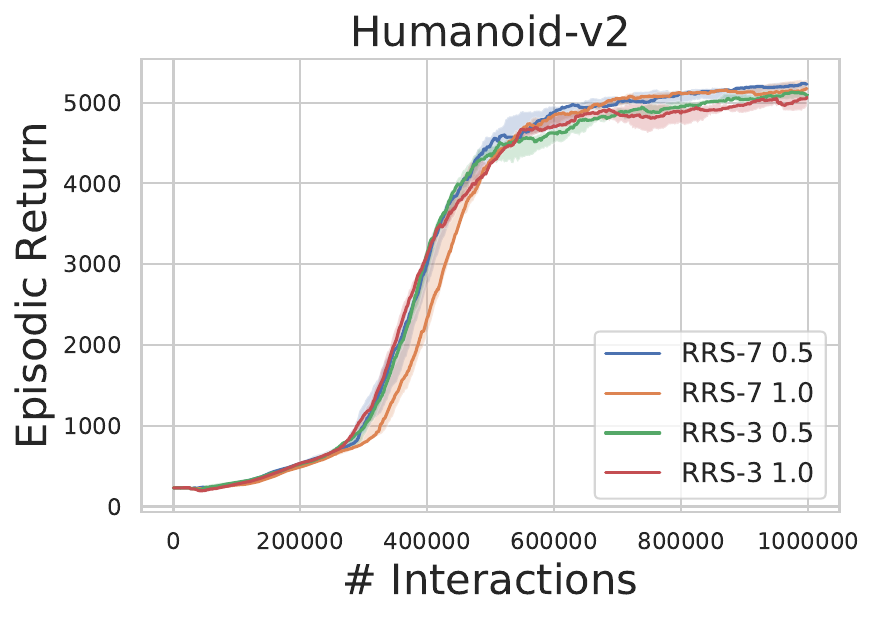}
\end{minipage}%
\caption{Performance with different reward shift constants and different number of $Q$-networks.}
\label{fig:ablation_cc}
\end{figure}

\paragraph{Ablation Studies}
We experiment with different number of $Q$-value networks as well as different choices of the random reward shifting ranges. Results are presented in Figure~\ref{fig:ablation_cc}. We denote RRS with $7$ reward shifting constants ( and therefore also $7$ $Q$-networks) as \textbf{RRS-7}, and denote RRS with $3$ reward shifting constants ( and therefore also $3$ $Q$-networks) as \textbf{RRS-3}. The constants following \textbf{RRS-3/RRS-7} are the ranges of those random constants. Specifically, we use $[-0.5, 0, 0.5]$ for the \textbf{RRS-3 0.5} settings, $[-1.0, 0, 1.0]$ for the \textbf{RRS-3 1.0} settings, $[-0.5, -0.33, -0.17, 0, 0.17, 0.33, 0.5]$ for the \textbf{RRS-7 0.5} settings and $[-1.0, -0.67, -0.33, 0, 0.33, 0.67, 1.0]$ for the \textbf{RRS-7 1.0} settings. According to the experimental results, RRS is not sensitive to hyper-parameters, showing the robustness of the proposed method. We believe further search for those hyper-parameters can further improve the learning efficiency, yet this is off the main scope of this work and therefore left for the future research.

\subsection{Random Network Distillation}
\label{appdx:RND}

\begin{figure}[h]
\centering
\begin{minipage}[htbp]{0.249\linewidth}
	\centering
	\includegraphics[width=0.8\linewidth]{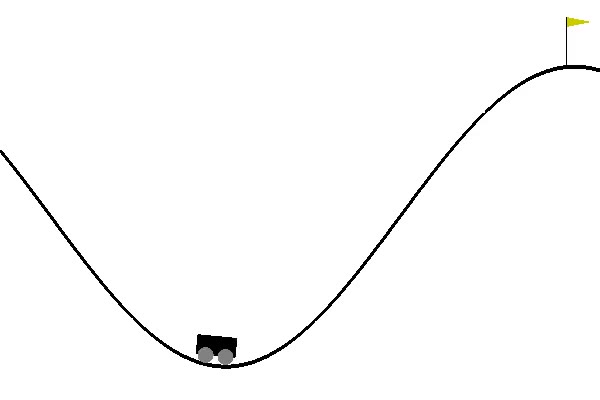}
\end{minipage}%
\begin{minipage}[left]{0.249\linewidth}
	\includegraphics[width=0.8\linewidth]{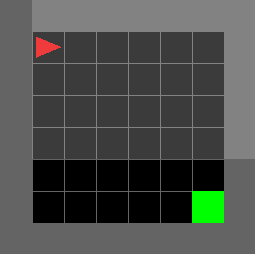}
\end{minipage}%
\begin{minipage}[htbp]{0.249\linewidth}
	\centering
	\includegraphics[width=0.8\linewidth]{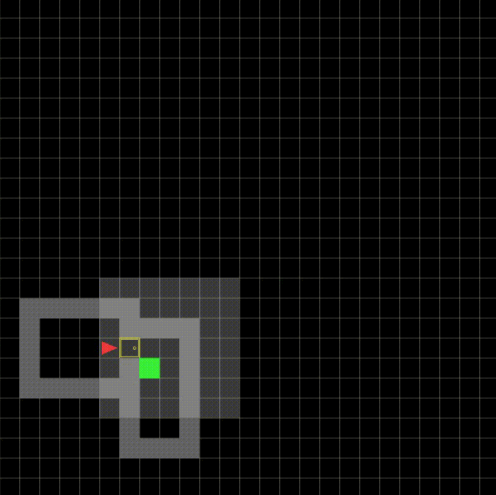}
\end{minipage}%
\begin{minipage}[htbp]{0.249\linewidth}
	\centering
	\includegraphics[width=0.8\linewidth]{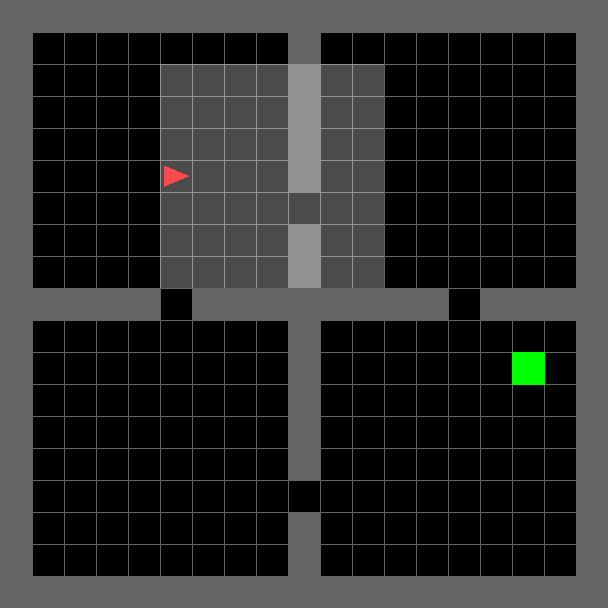}
\end{minipage}%
\caption{Examples of environments used in Section ~\ref{sec:curiosity_driven_exp}. The first figure shows the MountainCar-v0 environment where a car needs to accumulate potential energy to reach the flag, to receive a positive reward. The second figure shows the maze of the Empty-Random task with size of $6$, the third one shows the MultiRoom of level S2-N4, where there are $2$ rooms with size $4$, the last figure shows example of FourRoom task with size $17$. In our experiments, as we use the vanilla DQN as the baseline, which is not suitable for partial observable tasks, we use a smaller maze of size $7$ and $9$ to avoid further dependency on memories. In all tasks of the MiniGrid domain, the triangular red agent need to navigate to the green goal square, and the observable region is only a $7$x$7$ square the agent is facing to (i.e., the regions with shallower color in the last three figures).}
\label{fig:RNDtasks}
\end{figure}

\begin{figure}[t!]
\centering
\begin{minipage}[htbp]{0.33\linewidth}
	\centering
	\includegraphics[width=1\linewidth]{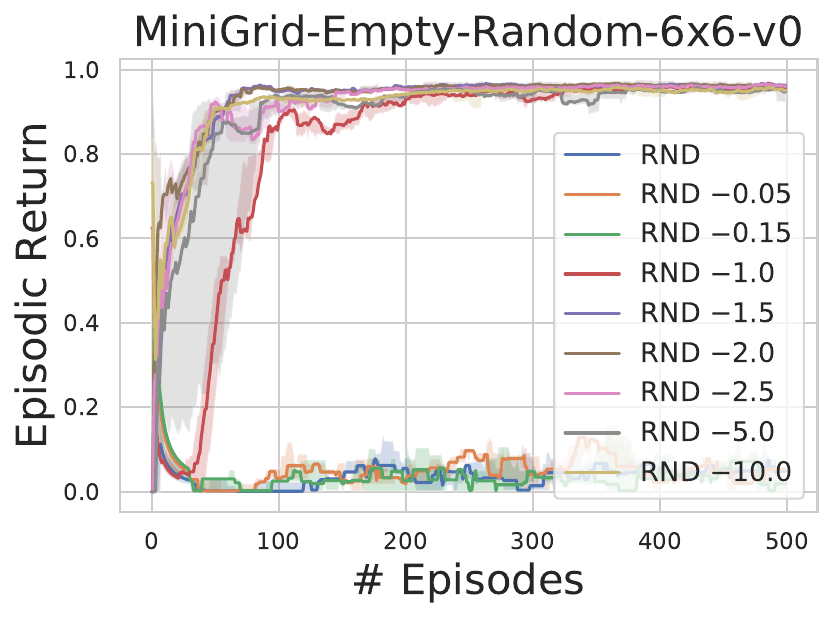}
\end{minipage}%
\begin{minipage}[htbp]{0.33\linewidth}
	\centering
	\includegraphics[width=1\linewidth]{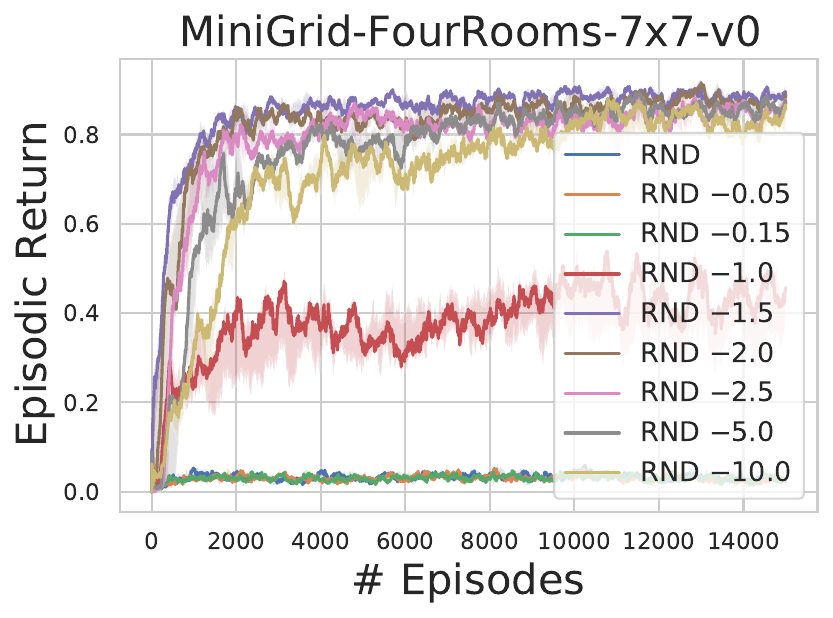}
\end{minipage}%
\begin{minipage}[htbp]{0.33\linewidth}
	\centering
	\includegraphics[width=1\linewidth]{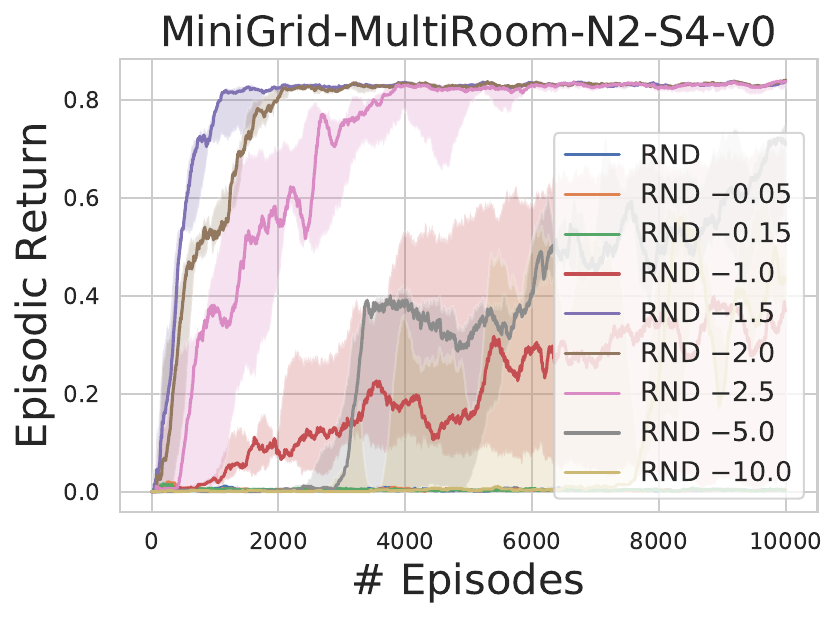}
\end{minipage}\\%
\begin{minipage}[htbp]{0.33\linewidth}
	\centering
	\includegraphics[width=1\linewidth]{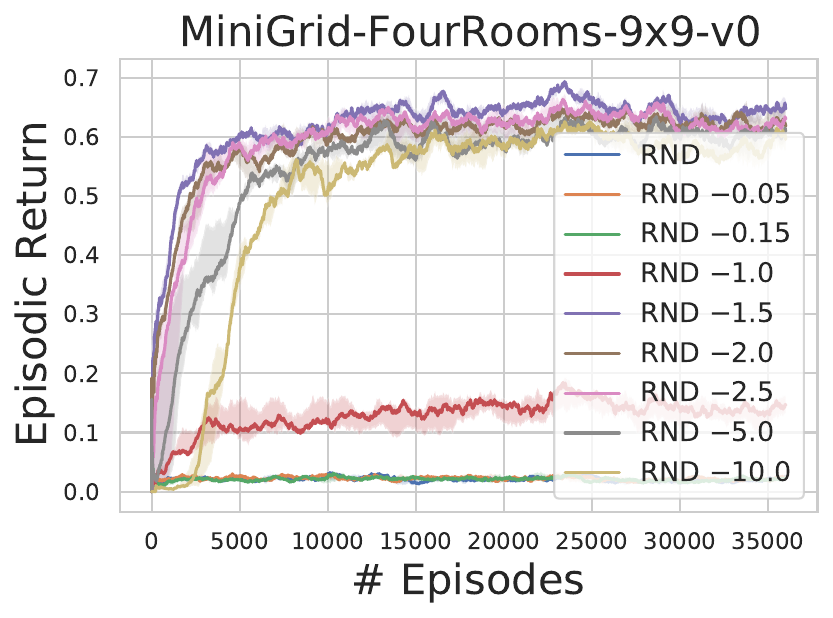}
\end{minipage}
\begin{minipage}[htbp]{0.5\linewidth}
	\centering
	\includegraphics[width=1\linewidth]{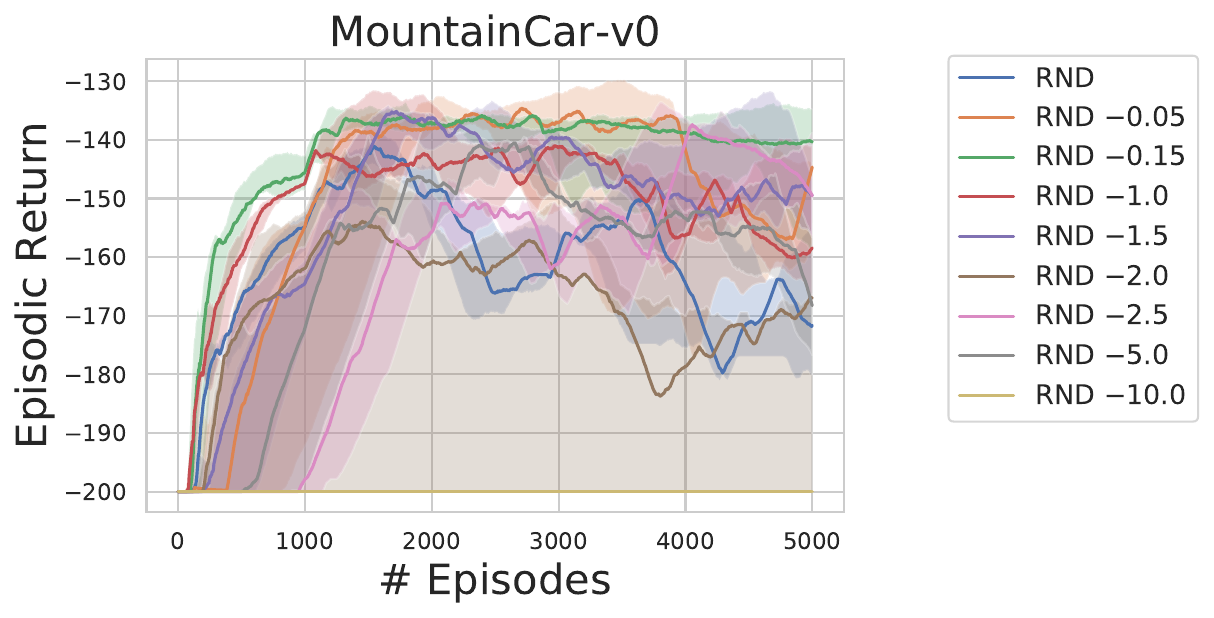}
\end{minipage}%
\caption{Performance with different reward shift constants in RND.}
\label{fig:ablation_rnd}
\end{figure}

\paragraph{Environments}
In this work, we experiment with five discrete (sparse reward) exploration tasks , namely the MountainCar-v0, and four navigation tasks of MiniGrid suite~\citep{gym_minigrid}, namely the task of Empty-Random, MultiRoom, and FourRooms, to verify our insight on improving RND for value-based curiosity-driven exploration. Figure~\ref{fig:RNDtasks} shows example of different tasks.
\paragraph{Hyper-Parameter Settings}
We use a $2$-layer NN with $64$ hidden units for $Q$-networks in DQN and set RND networks to be $3$-layer NN with $512$ hidden units. $\epsilon$-greedy exploration is applied to DQN with $\epsilon$ decays from $0.9$ to $0.05$ in the first 1/5 episodes. Size of replaybuffer is set to be $100000$.
\paragraph{Ablation Studies}
We experiment with different reward shifting constants in the discrete control settings. We use a relatively large range in choosing constants, i.e., $\{-0.05, -0.15, -1.0, -1.5, -2.0, -2.5, -5.0, -10.0\}$. Results are presented in Figure~\ref{fig:ablation_rnd}. In all experiments, using a moderate reward shifting constant like $\{-1.0, -1.5, -2.0, -2.5\}$ remarkably improves the learning efficiency. On the other hand, a too aggressive reward shifting will lead to too much curiosity exploration and hinder the learning efficiency in the limited number of interactions.

\newpage
\section{Additional Experiments}
\subsection{Necessity of Explorative Behaviors in Maze Tasks}
\begin{figure}[h]
\centering
\begin{minipage}[htbp]{0.5\linewidth}
	\centering
	\includegraphics[width=1\linewidth]{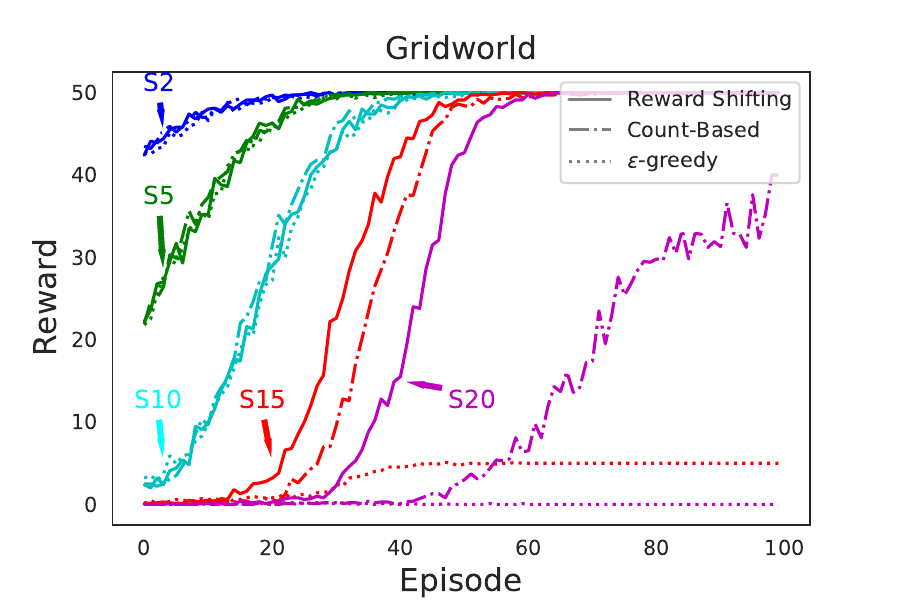}
\end{minipage}%
\caption{In deep-exploration tasks, reward shifting benefits exploration by optimistic initialization, while in easier tasks, reward shifting does not hinder exploitation, convergence efficiency, and the asymptotic performance.}
\label{fig:appdx_add_exp1}
\end{figure}

We demonstrate the benefits of reward shifting for deep-exploration tasks and the on-par performance of reward shifting for easy tasks that does not require deep-exploration. 
We experiment on the maze environment and change the size of maze to vary from 2 to 20, denoting as S2, S5, S10, S15, S20, separately. Results averaged over $8$ runs are reported in Figure~\ref{fig:appdx_add_exp1}. We find in easy tasks (S2, S5, S10), both count-based exploration and reward shifting perform similarly to the $\epsilon$-greedy exploration, while on challenging tasks (S15, S20), more explorative behavior encouraged by reward shifting and count-based exploration are important for efficient learning.

\subsection{Performance in Challenging Continuous Control Exploration Tasks}
\begin{figure}[h]
\centering
\begin{minipage}[htbp]{0.5\linewidth}
	\centering
	\includegraphics[width=1\linewidth]{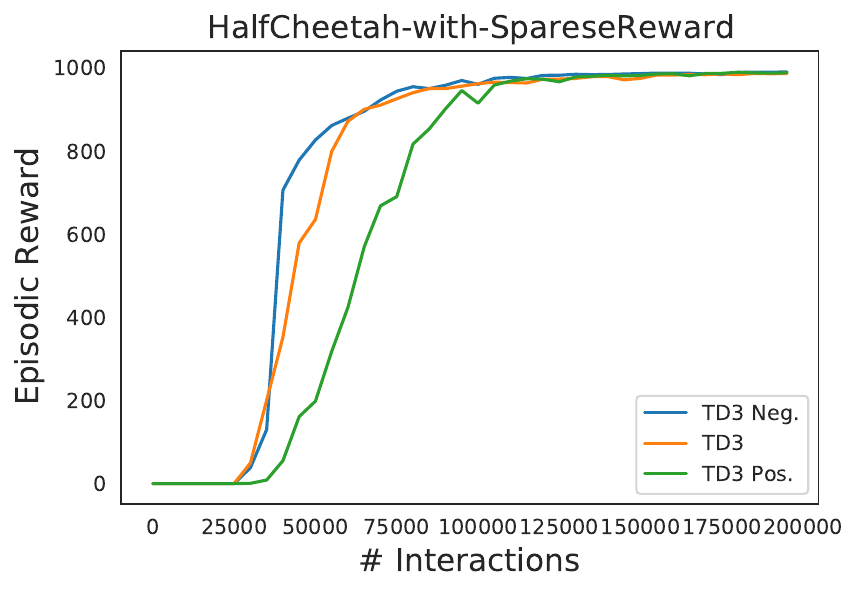}
\end{minipage}%
\begin{minipage}[htbp]{0.5\linewidth}
	\centering
	\includegraphics[width=1\linewidth]{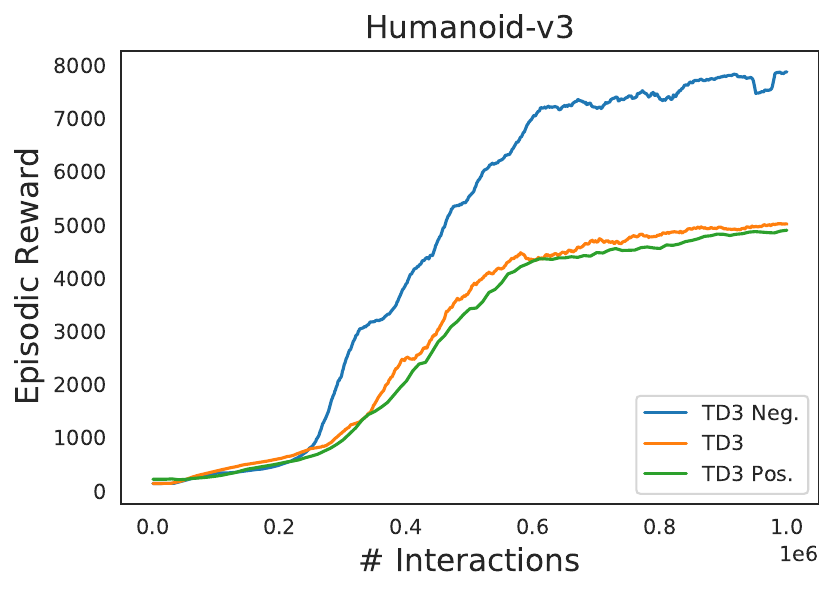}
\end{minipage}%
\caption{Experiments on two challenging continuous control tasks. Experiments are repeated with $8$ runs.}
\label{fig:appdx_add_exp2}
\end{figure}

Figure~\ref{fig:appdx_add_exp2} show experiments on (1) the HalfCheetah-SparseReward, where a reward of +1 is provided only when the forward movement of the halfcheetah is larger than 5 unit with regard to the timeframe; We note that this environment is different from the SparseHalfCheetah environment that first introduced in VIME~\cite{houthooft2016vime}, as we use different time frames. (We acknowledge and thank the anonymous reviewer FqMf for pointing out this difference.) and (2) the Humanoid, which is a high dimensional continuous control task in the MuJoCo locomotion suite; to verify the effectiveness of reward shifting in exploration. In the HalfCheetah-SparseReward environment, the maximum score is $1000$ for each episode. We use $-0.5$ as the negative reward shift for exploration and use $0.5$ for comparison. In Humanoid, the per-step reward is approximately $5$ in previous well-performing agents~\citep{fujimoto2018addressing,haarnoja2018soft}, and we hereby use negative shift $-5$ and use positive shift $5$, for comparison.

In HalfCheetah-SparseReward, we find a negative reward shift lead to more explorative behavior and improves the learning efficiency while a positive reward shift hinders the learning efficiency. 
In Humanoid, we find using a reward shift can drastically improve the asymptotic performance by ~$+60\%$, while learning with a positive reward shift retard the learning and converge to a lower performance. 

\subsection{Performance in Goal-Conditioned Continuous Control (Robotics) Suite}

\begin{figure}[h]
\centering
\begin{minipage}[htbp]{0.5\linewidth}
	\centering
	\includegraphics[width=1\linewidth]{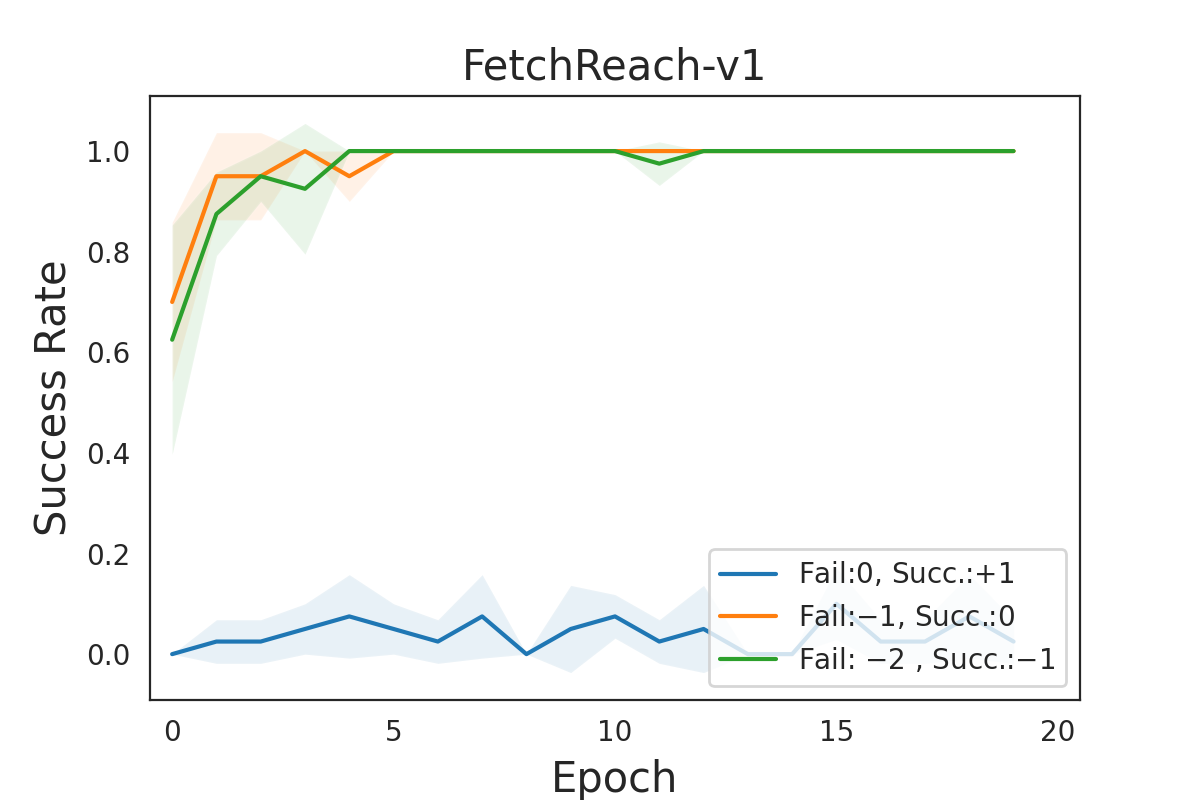}
\end{minipage}%
\begin{minipage}[htbp]{0.5\linewidth}
	\centering
	\includegraphics[width=1\linewidth]{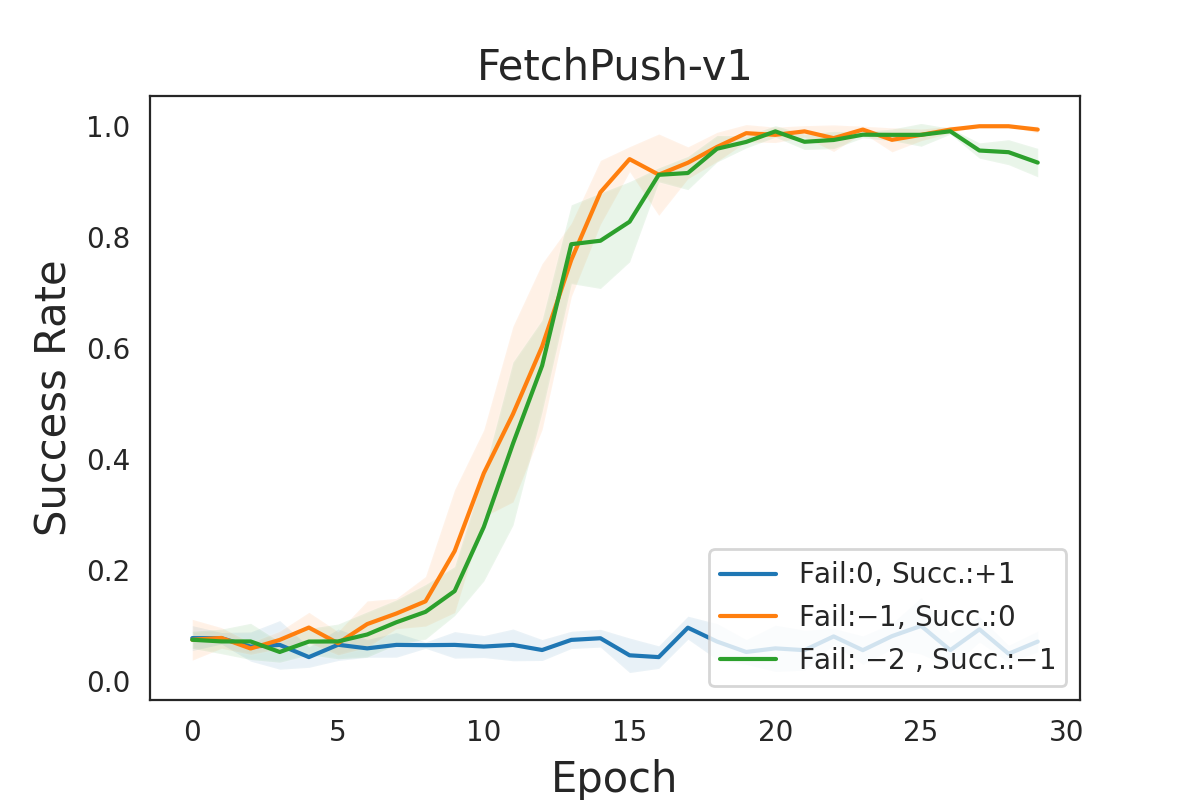}
\end{minipage}%
\caption{Experiments on two GCRL robotics tasks. Experiments are repeated with $4$ runs.}
\label{fig:fetch_robotics}
\end{figure}

To further address the reviewer's concern on the applicability of reward shifting on challenging continuous control exploration tasks, we benchmark reward shifting on the FetchRobotics suite~\citep{plappert2018multi} that is usurally considered to be challenging exploration task in GCRL literature~\citep{yang2022rethinking,andrychowicz2017hindsight}. Figure~\ref{fig:fetch_robotics} shows the results we get on FetchReach-v1 and FetchPush-v1 environment. We use HER~\citep{andrychowicz2017hindsight} as the backbone algorithms and vary the reward for failure and success in achieving the goals. In their default setting, reaching the goal will receive a reward of $0$, otherwise, the agent will receive $-1$ as punishment. In experiments, we find using a positive reward $+1$ in reaching the goals while using a trivial $0$ reward otherwise will drastically hinder the learning efficiency of HER. Similar empirical discovery has been reported in \citet{sun2019policy} in the PPO-based learners. This set of experiments verifies our key insight one more time that explorative behaviors emerge with a negatively shifted reward function, and a positive reward shift leads to conservative behavior.

To sum up, our key insight reveals the mechanism of how such an empirically verified heuristic design in GCRL works: a negative reward  $-1$ (also interpreted as cost) works in the same way as reward shifting to improve exploration.


\end{document}